\documentclass[10pt,journal,compsoc]{IEEEtran}
\ifCLASSOPTIONcompsoc
  \usepackage[nocompress]{cite}
\else
  \usepackage{cite}
\fi

\ifCLASSINFOpdf
   \usepackage[pdftex]{graphicx}
\else
\fi

\usepackage{amsmath,amssymb,multirow,amsthm}
\usepackage{tabularx,color,hyperref,multirow,booktabs}
\usepackage[table,xcdraw]{xcolor}

\usepackage[numbers]{natbib}
\usepackage{pifont}      
\ifCLASSOPTIONcompsoc
  \usepackage[caption=false,font=footnotesize,labelfont=sf,textfont=sf]{subfig}
\else
  \usepackage[caption=false,font=footnotesize]{subfig}
\fi

\usepackage[ruled,vlined]{algorithm2e}
\usepackage{algorithmic}

\usepackage[multiple]{footmisc}
\newcolumntype{Y}{>{\raggedright\arraybackslash}X}

\newtheorem{proposition}{Proposition} 
\newtheorem{theorem}{Theorem}

\providecommand{\eqdef}{\triangleq}
\begin{document}

% Title content
\title{Keep it Light! Simplifying Image Clustering via Text-Free Adapters}

\author{Yicen Li, Haitz Sáez de Ocáriz Borde, Anastasis Kratsios and Paul D. McNicholas~\IEEEmembership{Senior Member,~IEEE}%
\IEEEcompsocitemizethanks{\IEEEcompsocthanksitem Y.\ Li is a Ph.D.\ candidate, A.\  Kratsios is an Assistant Professor, and P.D.~McNicholas is the Canada Research Chair in Computational Statistics at the Department of Mathematics and Statistics, McMaster University, Hamilton, ON, Canada.\protect\\
E-mail: \{li2642,kratsioa,paulmc\}@mcmaster.ca}}%
\markboth{Li et al.: Keep it Light! Simplifying Image Clustering via Text-Free Adapters}%
{Li et al.: Keep it Light! Simplifying Image Clustering via Text-Free Adapters}

\IEEEtitleabstractindextext{%
\begin{abstract}
In the era of pre-trained models, effective classification can often be achieved using simple linear probing or lightweight readout layers. In contrast, many competitive clustering pipelines have a multi-modal design, leveraging large language models (LLMs) or other text encoders, and text-image pairs, which are often unavailable in real-world downstream applications. Additionally, such frameworks are generally complicated to train and require substantial computational resources, making widespread adoption challenging. In this work, we show that in deep clustering, competitive performance with more complex state-of-the-art methods can be achieved using a text-free and highly simplified training pipeline. In particular, our approach, Simple Clustering via Pre-trained models (SCP), trains only a small cluster head while leveraging pre-trained vision model feature representations and positive data pairs. Experiments on benchmark datasets, including CIFAR-10, CIFAR-20, CIFAR-100, STL-10, ImageNet-10, and ImageNet-Dogs, demonstrate that SCP achieves highly competitive performance. Furthermore, we provide a theoretical result explaining why, at least under ideal conditions, additional text-based embeddings may not be necessary to achieve strong clustering performance in vision.
\end{abstract}

}

\maketitle

\IEEEdisplaynontitleabstractindextext

\IEEEpeerreviewmaketitle

% Introduction and Overview
\IEEEraisesectionheading{\section{Introduction}\label{sec:intro}}
\IEEEPARstart{P}{owered} by the expressive capabilities of neural networks, deep clustering (DC) has redefined the state of the art (SOTA) in image clustering, outperforming classical algorithms by indisputable margins. Modern DC pipelines uncover subtle feature structures that substantially enhance the effectiveness of downstream learners. A highly non-exhaustive list of domains where DC-powered algorithms have reshaped the SOTA includes community detection~\citep{saeedi2018novel, yaroslavtsev2018massively}, anomaly detection~\citep{pang2021deep}, image segmentation~\citep{minaee2021image, liang2023clustsegclusteringuniversalsegmentation}, object detection~\citep{zhao2019object, kim2024garfieldgroupradiancefields}, and various medical applications~\citep{zhao2020joint}.

However, in addition to image encoders, many recent DC pipelines rely on \emph{massive} models—often involving large language models (LLMs)—that are unnecessarily compute-intensive for clustering tasks and introduce complicated training schemes. Unlike supervised learning methods, which typically require only a lightweight readout layer or linear probing for classification, current unimodal deep clustering frameworks frequently incorporate additional components such as support sets, feature projection layers, or teacher–student architectures. While these designs have achieved impressive performance, they also introduce extra hyperparameters, increase training complexity, and ultimately raise an important research question.

\subsection{Is SOTA performance achievable with a simple DC pipeline?}
As image clustering often serves as a foundational step for a variety of downstream tasks, a simple yet effective solution can thus provide broad utility to the community. Indeed, we show that competitive performance, SOTA and near-SOTA, can be achieved using only a small adapter modifying the features generated by a pre-trained text-free, image encoder, i.e.\ not requiring any additional input text from the user, LLMs, or any other variants of text encoders.  The result is a \textit{lightweight} and \textit{simple} end-to-end DC pipeline, which is computationally cheap enough to be run on a standard L4 GPU.  In parallel to how linear probing or read-out layers are deployed in classification tasks, we show that our adapter is easy to adopt in real-world applications, as it is simple and effective.

% nd does not require text-image pairs, which may not always be readily available in practical clustering applications. 

\subsection{Theoretical Motivation}

Text-image pairs may not always be readily available in practical clustering applications. The intuition behind why a text-free classifier should, at least theoretically, be able to match the power of a classifier using text embeddings together with the raw pixel data is rooted in our theoretical results. If $X$ represents a random image and $Z$ is its corresponding semantic description, which itself is a compressed representation of the information \textit{only in} the random image $X$, then any classification task $Y$ that depends on both the image $X$ and its textual information can ultimately be expressed as a function of the image $X$ alone.  
We denote the $\sigma$-algebra generated by the random variable $X$ by $\sigma(X)$.

\begin{proposition}[\textbf{Lossless Amortization Principle (LAR)}]
\label{prop:LosslessAmort}Fix $d,D,C\in \mathbb{N}$.
Let $X,Z$ be random variables respectively, taking values in $\mathbb{R}^d$ and in $\mathbb{R}^D$, both of which are defined on a common probability space $(\Omega,\mathcal{F},\mathbb{P})$, and suppose that $Z$ is $\sigma(X)$-measurable.  
For every $\{0,1,\dots,C-1\}$-valued random variable $Y$ on $(\Omega,\mathcal{F},\mathbb{P})$ if:
there exists a Borel measurable function $f:
(\mathbb{R}^{d+D},\mathcal{B}(\mathbb{R}^{d+D})) \to (\mathbb{R}^d,\mathcal{B}(\mathbb{R}^d))$ representing $Y$ by
\begin{equation}
\label{eq:uncompressed}
% \tag{Class}
\underbrace{
Y = 
f(X,Z)
}_{\text{Depends both on $X$ and $Z$}}
\end{equation}
then, there is a Borel map $F: (\mathbb{R}^d,\mathcal{B}(\mathbb{R}^d))\to 
(\mathbb{R}^n,\mathcal{B}(\mathbb{R}^n))$ providing the following lossless amortized representation
\begin{equation}
\label{eq:compression}
\tag{LAR}
% \underbrace{
    % \mathbb{E}[
    Y
    % |X] 
% }_{\text{Depends both on $X$ and $Z$}}
= 
\underbrace{
    F(X)
.
}_{\text{Only depends on $X$}}
\end{equation}
\end{proposition}

\begin{proof}[{Proof of Proposition~\ref{prop:LosslessAmort}}]
By the Doob–Dyknin lemma, see e.g.~\citep[Lemma 1.14]{KallenbergProbability_2021}, since $Z$ is $\sigma(X)$-measurable iff there exists some Borel measurable function $g:(\mathbb{R}^d,\mathcal{B}(\mathbb{R}^d))\to (\mathbb{R}^D,\mathcal{B}(\mathbb{R}^D))$ such that
\begin{equation}
\label{eq:setup_map}
Z = g(X).
\end{equation}
Define $F: (\mathbb{R}^d,\mathcal{B}(\mathbb{R}^d))\to (\mathbb{R}^n,\mathcal{B}(\mathbb{R}^n))$ by composition:
\begin{equation}
\label{eq:conclusion}
F(x) \eqdef f(x,g(x)).
\end{equation}
Then, by \eqref{eq:setup_map} and \eqref{eq:uncompressed},
\[
Y = f(X,Z) = f(X,g(X)) = F(X),
\]
which establishes the claim.
\end{proof}

Further, the proposition above seems to align with the \textit{Platonic Representation Hypothesis} proposed in~\citep{pmlr-v235-huh24a}, which suggests that latent representations in deep networks converge despite being trained on distinct data modalities. This supports our claim that, as long as the backbone vision model provides sufficiently good latent representations, strong performance can be achieved through a simple DC pipeline without relying on additional modality inputs.

We acknowledge that, generally speaking, incorporating multimodal representations and additional information in our network tends to yield better empirical performance. Rather than disputing this, we deliberately refrain from using such frameworks, and instead aim to explore whether, in principle and under ideal conditions, clustering could be achieved using the simplest model possible, without multimodal data collection.

Our LAR principle (Proposition~\ref{prop:LosslessAmort}) shows that any semantically-powered image classifier may be realized by a text-free classifier ``$F$''.  However, \textit{can this theoretical classifier be practically realized, at-least approximately?}
Our main theoretical result guarantees that the theoretical text-free classifiers can indeed be approximately implemented by an MLP with a $\operatorname{softmax}$ output activation.  Our guarantee holds for multiclass MLP classifiers using any standard activation function such as $\operatorname{ReLU}$, $\operatorname{Swish}$, or $\operatorname{softplus}$.

\begin{theorem}[\textbf{Text-Free DC is Powerful Enough}]
\label{thrm:DCplus}
In the setting of Proposition~\ref{prop:LosslessAmort}, let $\rho:\mathbb{R}\to \mathbb{R}$ be an activation function with at least one point of continuous and non-zero differentiability.  
Then, for every $0<\varepsilon\le 1$, there is an MLP $\hat{F}:\mathbb{R}^d\to \mathbb{R}^C$, with activation function $\rho$, satisfying
\begin{equation}
        \big|
            \underbrace{
                f(X,Z)
            }_{\text{text-dependent classes}}
            -
            \underbrace{
                \,\,
                \operatorname{softmax}\circ \hat{F}(X)
            }_{\text{text-free deep learner}}
        \big|
    <
        \varepsilon
\end{equation}
with probability at least $1-\varepsilon$.
\end{theorem}

\begin{proof}[{Proof of Theorem~\ref{thrm:DCplus}}]
Let $\mu \eqdef X_{\#}\mathbb{P}$ denote the law of $X$.  
By \cite[Theorem~13.6]{KlenkeBook_2020}, $\mu$ is a Radon measure.  
Since $\mathbb{R}^d$ is second-countable and locally compact,  
Lusin's theorem \cite[Exercise~13.1.3]{KlenkeBook_2020} yields,  
for every $\varepsilon \in (0,1]$, a compact set 
$\mathcal{K}_\varepsilon \subset \mathbb{R}^d$ such that
\begin{equation}
\label{eq:High_Prob_Lusin}
\mu(\mathcal{K}_{\varepsilon}) \ge 1 - \varepsilon,
\end{equation}
and $F$ is continuous on $\mathcal{K}_{\varepsilon}$.

Now, let $\rho$ satisfy \cite[Assumption~1]{kratsios2022universal}
(as in \cite{kidger2020universal}), and recall that the softmax activation
satisfies \cite[Assumption~8]{kratsios2022universal}.  
Then, the universal approximation theorem
\cite[Theorem~37(ii)]{kratsios2022universal} ensures the existence of an
MLP $\hat{F}:\mathbb{R}^d \to \mathbb{R}^C$ such that
\begin{equation}
\label{eq:uniform_approx}
\sup_{x \in \mathcal{K}_{\varepsilon}}
\big|
F(x) - \operatorname{softmax} \circ \hat{F}(x)
\big|
< \varepsilon.
\end{equation}

Define
\[
B \eqdef
\bigl\{
x \in \mathbb{R}^d :\,
|F(x) - \operatorname{softmax} \circ \hat{F}(x)|
\ge \varepsilon
\bigr\}.
\]
Then $\mathcal{K}_{\varepsilon} \subseteq \mathbb{R}^d \setminus B$.  
Let $L(x) = |F(x) - \operatorname{softmax} \circ \hat{F}(x)|$.  
Since $L$ is measurable, so is 
$B = L^{-1}([\varepsilon,\infty))$.

Combining \eqref{eq:High_Prob_Lusin} and \eqref{eq:uniform_approx}, we get

\begin{center}
\resizebox{\columnwidth}{!}{$
\begin{aligned}
\mathbb{P}\!\Big(
  |F(X)
  - \operatorname{softmax}\!\circ\!\hat{F}(X)|
  < \varepsilon
\Big)
&= \mu\!\Big(
  |F(x)
  - \operatorname{softmax}\!\circ\!\hat{F}(x)|
  < \varepsilon
\Big) \\
&= \mu(\mathbb{R}^d \setminus B) \\
&\ge \mu(\mathcal{K}_{\varepsilon}) \\
&\ge 1 - \varepsilon.
\end{aligned}
$}
\end{center}
\label{eq:finishme}

Finally, by Proposition~\ref{prop:LosslessAmort} and \eqref{eq:uncompressed},
\begin{equation}
\label{eq:rep_me}
F(X) = f(X, Z).
\end{equation}

Substituting \eqref{eq:rep_me} into \eqref{eq:finishme} gives
\begin{align}
\mathbb{P}\!\Big(
  |f(X, Z)
  - \operatorname{softmax}\!\circ\!\hat{F}(X)|
  < \varepsilon
\Big)
&\ge 1 - \varepsilon.
\label{eq:finishme2}
\end{align}
which concludes the proof.
\end{proof}

\subsection{Simplifying Contemporary Deep Clustering}

Modern clustering pipelines, e.g. Van et al.\ and Ding et al. \citep{van2020scan,ding2023unsupervised} usually involve: \textbf{(i)} learning powerful initial latent representations through self-supervised techniques, e.g.\ joint-embedding approaches \citep{chen2020simple,he2020momentum}, and \textbf{(ii)} gradually refining the representation and clustering membership by minimizing an objective function \citep{caron2019deepclusteringunsupervisedlearning}. 

\paragraph{Representation Learning in Deep Clustering} 
The emergence of large-scale pre-trained models optimized according to a contrastive learning objective using both natural language such as CLIP~\citep{radford2021learning} and purely image-based encoders like DINO~\citep{caron2021emerging,oquab2024dinov2learningrobustvisual}, has proven highly effective in capturing rich, general-purpose feature representations for a variety of downstream tasks. 
Most SOTA-achieving DC pipelines leverage these powerful representations alongside text-embedding~\citep{li2023image,cai2023semantic} (the $Z$ in Theorem~\ref{thrm:DCplus}) for effective latent feature alignment.  However, it would involve extra computations and we will see in Section~\ref{sec:Experiments}, these text-based representations seem to contain redundant features, given that comparable performance can be achieved with a unimodal framework. As we have already mentioned, being able to eliminate the need for text would make downstream applications, particularly in data-scarce regimes, much more accessible to the end user.

\paragraph{Self-Supervised Training}
Typically, modern DC pipelines are trained using self-supervised methods by optimizing contrastive losses~\citep{oord2018representation}. While contrastive methods rely on both positive and negative pairs, other approaches like BYOL~\citep{grill2020bootstrap} have demonstrated that it is possible to learn effective representations without negative pairs by simply maximizing the agreement between two views of the same data. This suggests that maximizing similarities of positive pairs alone can yield high-quality representations. We will follow a similar approach in this work too, see Section~\ref{sec:Our Method}.

\paragraph{From Theory to Practice} Even if the model is approximately optimal and its parameters are perfectly adjusted, it is not clear that any real-world unsupervised training procedure can achieve the theoretical expressivity guaranteed by Theorem~\ref{thrm:DCplus}. Hence, we would like to empirically answer whether \textit{a simple DC pipelines can rival both multimodal methods and its unimodal competitors}. We would like to emphasize that obtaining powerful DC performance, with the additional requirement that our DC pipeline is simple, and text-free, is highly nontrivial since even the current SOTA in deep clustering exhibits a \textit{substantial} performance gap with respect to supervised image classification models.  Indeed, supervised methods typically achieve accuracies well-above $90\%$ on image datasets, while unsupervised clustering approaches rarely surpassed 50\% clustering accuracy on CIFAR-20, and many fall below $50\%$~\citep{dang2021nearest,li2021contrastive,shen2021you,niu2022spice}. This is because most unsupervised clustering pipelines struggle with complex natural images due to high intra-class variability.

\paragraph{Contributions} To address the challenges discussed so far, we propose Simple Clustering via Pre-trained models (SCP), particularly in scenarios where text information or complex frameworks are not accessible. Much like linear probing in classification, SCP provides a simple yet powerful way to leverage pre-trained visual features for accessible deep clustering. Our main contributions are as follows:

\begin{itemize}
\item \textbf{Simplicity:}
Our method yields competitive results while maintaining a simple yet effective architecture, requiring no additional feature layers, support sets, teacher-student networks, text information, or exponential moving averages during training. 

\item \textbf{Applicability:}
Our method seamlessly integrates the powerful image encoder from various pre-trained models such as CLIP and DINO into a self-supervised clustering framework. We show that this text-free adapter achieves highly competitive clustering performance on benchmark datasets such as CIFAR-10 \citep{krizhevsky2009learning}, CIFAR-20 \citep{krizhevsky2009learning} and STL-10 \citep{krizhevsky2009learning} to current SOTA.

\item \textbf{Principled Approach:} 
Our text-free approach is principled, based on the approximate representation result previously presented in Theorem~\ref{thrm:DCplus}, which guarantees that text-free embeddings are theoretically enough.
\end{itemize}

\section{Related Work}
In this section, we provide a broad overview of self-supervised learning research that has inspired our work, along with recent trends in image clustering using pre-trained models.

\subsection{Self-Supervised Learning}
Self-supervised learning learns representations from data without explicit labels. The objective is to create a representation space where positive pairs are closer together, while negative pairs are pushed farther apart \citep{geiping2023cookbook}.

SimCLR \citep{chen2020simple} uses data augmentations, such as flipping and colour jittering, to create positive and negative pairs for optimizing objectives. It also introduces a projection head that maps embeddings into a space where contrastive loss is applied. BYOL \citep{grill2020bootstrap} shows that high-quality representations can be learned by simply maximizing agreement between two augmented views of the same input, without requiring negative pairs. Building on these advancements, SimSiam \citep{chen2020exploringsimplesiameserepresentation} eliminates the need for both negative pairs and momentum encoders by introducing a stop-gradient operation, which effectively prevents representational collapse. Inspired by these methods, we adopt similar ideas to develop a simple and effective self-supervised framework for image clustering.

\subsection{Pre-trained Models in Vision} 
Building on advances in self-supervised learning, CLIP \citep{radford2021learning} introduced a paradigm of contrastive pre-training that aligns images with corresponding textual descriptions. This approach enables broad task generalization without task-specific fine-tuning. DINO \citep{caron2021emerging}, which stands for self-distillation with no labels, demonstrates a self-supervised method for optimizing a student network from a teacher network based on vision input data only. The scalability of CLIP has been further validated by openCLIP \citep{Cherti_2023}, which extended CLIP using the larger Vision Transformer models \citep{dosovitskiy2020image}. Similarly, models such as DINO~\citep{9709990} and DINOv2~\citep{oquab2024dinov2learningrobustvisual} are capable of processing visual data and mapping it to high-quality latent representations.

% One of the key advantages of pre-trained models like CLIP is their ability to eliminate the need for training models from scratch for downstream tasks, significantly reducing computational costs and time. Instead of training a self-supervised neural network from the ground up, pre-trained models provide high-quality feature representations out of the box, leading to faster experimentation and improved performance on a variety of tasks. 
% The scalability of CLIP has been further validated by openCLIP \cite{Cherti_2023}, which extended CLIP using the larger Vision Transformer models \cite{dosovitskiy2020image}. Similarly, models such as DINO~\cite{9709990} and DINOv2~\cite{oquab2024dinov2learningrobustvisual} are capable of processing visual data and mapping it to high-quality latent representations.

\subsection{Image Clustering via Pre-trained Models}
To address the challenges of scaling to modern image datasets, methods such as NMCE \citep{li2022neural} and MLC \citep{deng2023acp} have integrated deep learning with manifold clustering using the minimum coding rate principle \citep{Arthur_Vassilvitskii_2007}. Building on this idea, CPP \citep{chu2024image} further refines CLIP features and estimates the optimal number of clusters when unknown. TEMI \citep{adaloglou2023exploring} improves clustering by leveraging associations between image features, introducing a variant of pointwise mutual information with instance weighting. Unlike our approach, TEMI utilizes a nearest-neighbors set and an exponential moving average for parameter optimization.

SIC \citep{cai2023semantic} leverages multi-modality by mapping images to a semantic space and generating pseudo-labels based on image-semantic relationships. More recently, TAC~\citep{li2023image} utilizes the textual semantics of WordNet~\citep{miller1995wordnet} to enhance image clustering by selecting and retrieving nouns that best distinguish the images, facilitating collaboration between text and image modalities through mutual cross-modal neighborhood distillation. 

Current pre-trained approaches often rely on text or complex architectures to ensure consistency, motivating us to develop a simple yet effective pipeline for image clustering. Our method requires only a lightweight clustering head and basic data augmentations, demonstrating competitiveness, simplicity, and applicability among recent models.

\section{Our Method}
\label{sec:Our Method}

Recent methods such as CC \citep{li2021contrastive} and CPP \citep{chu2024image} decouple the latent space into clustering and feature spaces. SIC \citep{cai2023semantic} and TAC \citep{li2023image} utilize text information, and TEMI \citep{adaloglou2023exploring} employs self-distillation networks to enhance clustering. Despite the promising results, these methods involve additional components and complicated training as suggested by the table \ref{tab:hyperparam_summary} below.
\begin{table}[ht]
\caption{Summary of tunable hyperparameters and their concise roles in recent deep clustering methods. Fixed hyperparameters are omitted.}
\label{tab:hyperparam_summary}
\centering
\footnotesize
\setlength{\tabcolsep}{4pt}
\begin{tabular}{l p{0.45\linewidth}} 
\toprule
\textbf{Methods} & \textbf{Hyperparameters} \\
\midrule
CPP \citep{chu2024image} & $K_{\max}$,  $\gamma$, $\varepsilon$\\
& max \#  clusters; entropy etc. \\
\addlinespace[0.1em]
TEMI \citep{adaloglou2023exploring} & $C$, $\beta$, $k$, $m$ \\
& \# clusters; PMI trade-off etc. \\
\addlinespace[0.1em]
TAC \citep{li2023image} & $K$, $\tilde N$, $\hat{\tau}$, $\hat{N}$, $\alpha$ \\
& \# clusters; center size etc. \\
\addlinespace[0.1em]
SIC \citep{cai2023semantic} & $c$, $\gamma_u$, $\gamma_r$, $\xi_c$, $\xi_a$, $k$, $\lambda$, $\beta$, $\tau_m$ \\
& \# clusters; noun pruning etc. \\
\addlinespace[0.1em]
\textbf{SCP (Ours)} & \textbf{$K$, $\alpha$} \\
& \textbf{\# clusters; loss weight} \\
\bottomrule
\end{tabular}
\end{table}

\begin{figure}[ht]
    \centering
    \includegraphics[width=\columnwidth]{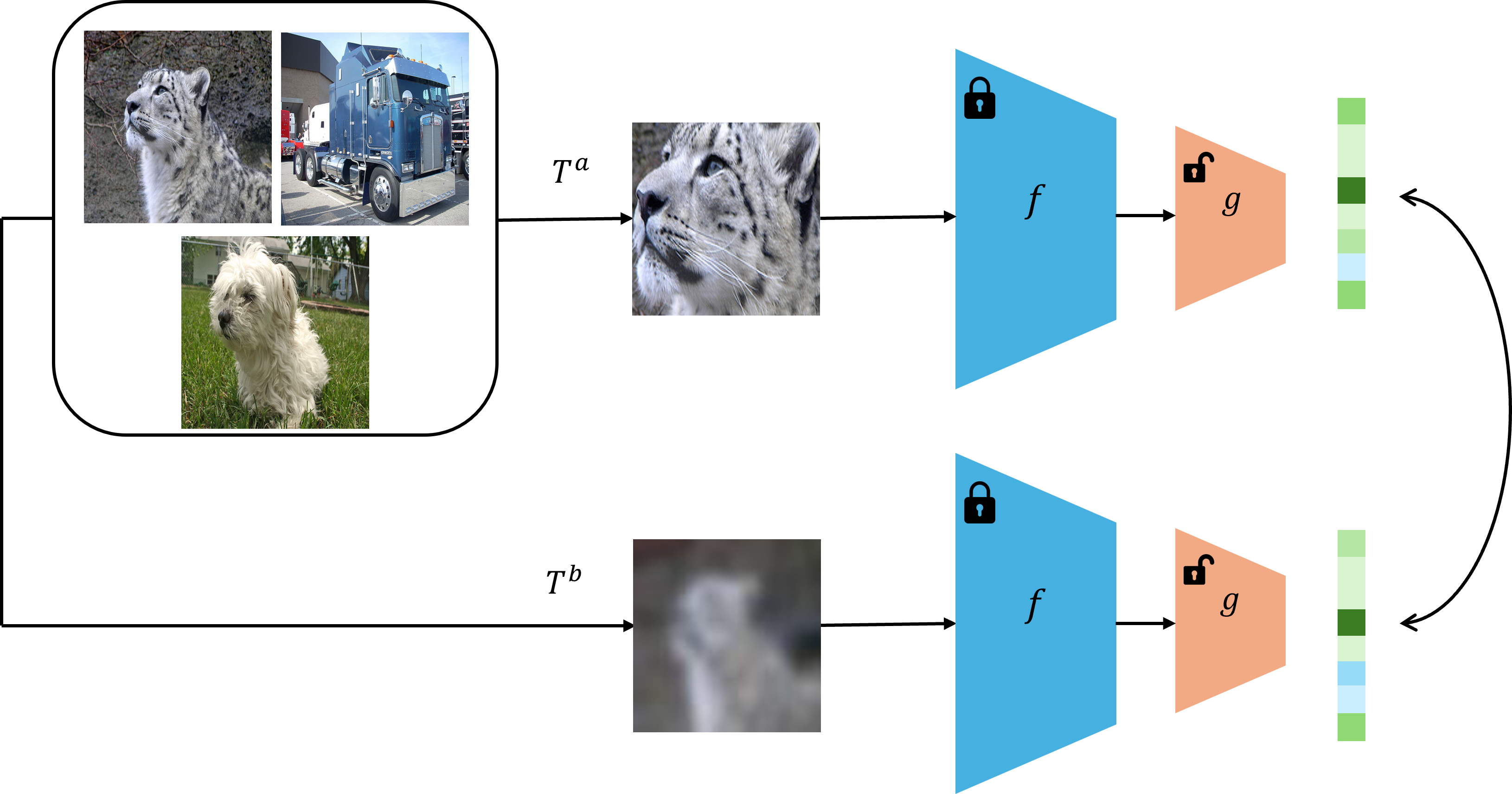}
    \caption{%
    A overall pipeline for SCP. During training, two augmented views \(T^a\) and \(T^b\) of an image are generated from the dataset and processed by a frozen feature extractor \(f\) and a trainable cluster head \(g\) (a five-layer MLP). The objective is to minimize the cross-entropy loss between the outputs of the cluster head \(g\) for the two augmented views.
    }
    \label{fig:AE}
\end{figure}

In contrast, in supervised learning, a lot of work perform classification only need a read-out layer or a simple linear-probing. Our approach try to keep the similar structure in the clustering as well and remains simple and efficient without relying on extra complexity as much as possible. As shown in Fig.~\ref{fig:AE}, SCP involves only two tunable hyperparameters and comprises two components: a pre-trained frozen backbone for pair construction, denoted as $f(.)$, and a trainable cluster head $g(.)$.
% \begin{wrapfigure}{r}{0.5\textwidth}
%     \centering
%     % \vspace{-1em}
%     \includegraphics[width=\linewidth]{images/scp.png}
%     % \vspace{-1em}
%     \caption{%
%     The overall pipeline of SCP. During training, two augmented views \(T^a\) and \(T^b\) of an image are generated from the dataset and processed by a frozen feature extractor \(f\) and a trainable cluster head \(g\) (a five-layer MLP). The objective is to minimize the cross-entropy loss between the outputs of \(g\) for the two augmented views.
%     }
%     \label{fig:AE}
% \end{wrapfigure}
Briefly, SCP performs data augmentations and extracts features from the augmented images using pre-trained models. The cluster head then projects these features into a cluster space, where the dimension equals the number of clusters. After training, outputs in the cluster space provide the soft assignments for clustering.

\subsection{Pair Construction Backbone}

The success of BYOL demonstrates that we can maximize the similarities of positive pairs without negative ones. In SCP, the positive pairs consist of samples augmented from the same instance. Given a data instance \(x_i\), we apply two stochastic transformations \(T^a\) and \(T^b\), independently selected from the augmentation family \(\mathcal{T}\). This produces two correlated views: \(x_i^a = T^a(x_i)\) and \(x_i^b = T^b(x_i)\). 

Although comprehensive augmentation strategies could enhance downstream performance, we adopt only two simple and commonly used augmentations in our work: \texttt{RandomCrop} and \texttt{GaussianBlur}. This choice aligns with the preprocessing techniques used in training pre-trained models, ensuring compatibility with their learned representations. \texttt{RandomCrop} randomly crops the image to a specified size, and \texttt{GaussianBlur} applies a Gaussian filter to blur the image. For each image, these two augmentations are applied independently, each with a 50\% probability. We then use a pre-trained model \(f(\cdot)\), such as CLIP, to extract features from the augmented images: $h_i^a = f(x_i^a)$ and $h_i^b = f(x_i^b).$

\subsection{Cluster Head}
Following the “label as representation” concept \citep{li2021contrastive}, when a data sample is projected into a space whose dimensionality matches the number of clusters \(K\), the \(k\)-th component of its feature vector (after applying a $\operatorname{softmax}$  function) can be interpreted as the probability that the sample belongs to the \(k\)-th cluster. We employ a five-layer non-linear MLP as the clustering head \(g(\cdot)\), producing a  \(K\)-dimensional feature that is normalized with a $\operatorname{softmax}$  over the dimension of the cluster.
\[
y_i^a = g(h_i^a), 
\quad 
y_i^b = g(h_i^b).
\]
Hence, \(y_i^a\) and \(y_i^b\) are both \(K\)-dimensional vectors, whose components \(y_{i,k}^a\) and \(y_{i,k}^b\) indicate the probability of assigning the \(i\)-th sample to the \(k\)-th cluster. 

% Formally, let \(Y^a, Y^b \in \mathbb{R}^{N \times K}\) be the outputs of the clustering head for all samples. Then, we have the following matrices:
% \[
% % Y^a = \begin{bmatrix}
% % y^a_1 \\
% % y^a_2 \\
% % \vdots \\
% % y^a_N
% % \end{bmatrix}
% % \quad
% % Y^b = \begin{bmatrix}
% % y^b_1 \\
% % y^b_2 \\
% % \vdots \\
% % y^b_N
% % \end{bmatrix}.
% Y^a = \begin{bmatrix}
% y^i_1 \\
% % y^a_2 \\
% \vdots \\
% y^a_N
% \end{bmatrix}
% \quad
% Y^b = \begin{bmatrix}
% y^b_1 \\
% % y^b_2 \\
% \vdots \\
% y^b_N
% \end{bmatrix}.
% \]

To maximize row-wise similarity, we adopt the following cross-entropy loss function instead of the commonly used InfoNCE loss \citep{oord2018representation}, as SCP only have positive pairs that should share similar soft assignments:
\begin{equation}
L_e = - \sum_{i=1}^{N} \sum_{k=1}^{K} y^{a}_{i,k} \log y^{b}_{i,k}.
\end{equation}
Inspired by the effective regularizations in previous works such as TAC \citep{li2023image}, we further introduce the following confidence loss to make the soft labels \( y^{a}_i \) and \( y^{b}_i \) more confident, approaching one-hot vectors:
\begin{equation}
L_{\text{con}} = - \log \sum_{i=1}^N {y^a_i}^\top y^b_i.
\end{equation}
This loss ensures that the cluster head assigns higher probabilities to its top predicted clusters, thereby increasing confidence in the assignments. 

In addition, following TAC \citep{li2023image}, we introduce an entropy term \( H(Y) \) to prevent model collapse, defined as follows:
\begin{equation}
H(Y) = - \sum_{k=1}^{K} \left[ P^{a}_k \log P^{a}_k + P^{b}_k \log P^{b}_k \right],
\end{equation}
where $P^{a}_k = \frac{1}{N} \sum_{i=1}^{N} y^{a}_{i,k}, \quad P^{b}_k = \frac{1}{N} \sum_{i=1}^{N} y^{b}_{i,k}.$ And $H(Y)$ encourages uniform soft assignments across clusters, thereby mitigating the issue of empty clusters.  Hence, we define the overall objective function of SCP as
\begin{equation}
L_{\text{clu}} = L_e + L_{\text{con}} - \alpha H(Y),
\end{equation}
where the balancing weight $\alpha$ modulates the influence of $H(Y)$, especially when the number of clusters is large. By maximizing consistency between different augmented views with regularizations, SCP effectively prevents trivial solutions and achieves competitive performance. We also provide Algorithm~\ref{alg:CAC} to explain our pipeline.

\begin{algorithm}[ht]
\label{alg:CAC}
\caption{Simple Clustering via Pre-trained Models (SCP)}
\KwIn{$\mathcal{X}=\{x_i\}_{i=1}^{N}$, pre-trained model $f(\cdot)$, clusters $K$, batch size $B$, loss weight $\alpha$}
\KwOut{Soft assignments $y_i=g(h_i)$ for each $x_i$}
Initialize cluster head $g(\cdot)$\;
\For{each epoch}{
  \For{each mini-batch $\{x_i\}_{i=1}^{B}$}{
    \tcc{Pair Construction}
    \For{each $x_i$ in mini-batch}{
      $x_i^a=T^a(x_i)$, $x_i^b=T^b(x_i)$\;
      $h_i^a=f(x_i^a)$, $h_i^b=f(x_i^b)$\;
    }
    \tcc{Cluster Space Encoding}
    \For{each $h_i^a, h_i^b$}{
      $y_i^a=g(h_i^a)$, $y_i^b=g(h_i^b)$\;
    }
    \tcc{Compute Losses}
    $L_{\mathrm{clu}}=L_e+L_{\mathrm{con}}-\alpha H(Y)$\;
    Update parameters of $g(\cdot)$ by minimizing $L_{\mathrm{clu}}$\;
  }
}
\end{algorithm}

% We provide algorithm \ref{alg:CAC} to explain our pipeline.

% \begin{algorithm}[h]
% \caption{Simple Clustering via Pre-trained Models}
% \label{alg:CAC}
% \begin{algorithmic}[1]
%     \STATE {\bf Input:} Dataset $\mathcal{X} = \{ x_i \}_{i=1}^{N}$, Pre-trained model $f(\cdot)$, number of clusters $K$, batch size $B$, loss weight $\alpha$
%     \STATE {\bf Initialize:} cluster head $g(\cdot)$ 
%     \FOR{epoch = 1 to maxEpoch}
%         \FOR{each mini-batch $\{x_i\}_{i=1}^B$}
%             \STATE {\bf Pair Construction:}
%             \FOR{each data instance $x_i$ in the mini-batch}
%                 \STATE Apply stochastic transformations $T^a$, $T^b$ to obtain:
%                 \STATE \quad $x_i^a = T^a(x_i)$, \quad $x_i^b = T^b(x_i)$
%                 \STATE Extract features using pre-trained model:
%                 \STATE \quad $h_i^a = f(x_i^a)$, \quad $h_i^b = f(x_i^b)$
%             \ENDFOR
%             \STATE {\bf Cluster Space Encoding:}
%             \FOR{each feature $h_i^a$, $h_i^b$}
%                 \STATE Compute soft assignments: 
%                 \STATE \quad $y_i^a = g(h_i^a)$, \quad $y_i^b = g(h_i^b)$
%             \ENDFOR
%             \STATE {\bf Compute Losses:}
%             \STATE Compute total clustering loss:
%             \STATE \quad $L_{\text{clu}} = L_e + L_{\text{con}} - \alpha\,H(Y)$
%             \STATE {\bf Update} cluster head $g(\cdot)$ by minimizing $L_{\text{clu}}$
%         \ENDFOR
%     \ENDFOR
%     \STATE \textbf{Return} soft assignments $y_i = g(h_i)$ for each $x_i \in \mathcal{X}$
% \end{algorithmic}
% \end{algorithm}

\section{Experiments}
\label{sec:Experiments}
In this section, we evaluate the proposed SCP on eight widely used image clustering datasets. A series of initial quantitative and qualitative comparisons, ablation studies, and hyper-parameter analyses will be carried out to investigate the effectiveness and robustness of the method.

\subsection{Experimental Setup}
We first introduce the datasets and metrics used for evaluation and then provide the implementation details of SCP.

\subsubsection{Datasets}
To evaluate the performance of SCP, we apply it to six widely used image clustering datasets: STL-10 \citep{coates2011analysis}, CIFAR-10 \citep{krizhevsky2009learning}, CIFAR-20 \citep{krizhevsky2009learning}, CIFAR-100 \citep{krizhevsky2009learning}, ImageNet-10 \citep{chang2017deep}, and ImageNet-Dogs \citep{chang2017deep}, which is a subset of ImageNet-1k \citep{deng2009imagenet}. In addition, we include two more challenging datasets, DTD \citep{cimpoi2013describingtextureswild} and UCF-101 \citep{soomro2012ucf101dataset101human}. The brief information of all datasets used in our main evaluation is summarized in the following table:

\begin{table}[t]
\caption{Characteristics of the benchmark datasets used in our evaluation.}
\label{tab:datasets}
\centering
\footnotesize
\resizebox{\columnwidth}{!}{%
\begin{tabular}{l c c c c}
\toprule
Dataset & Split (Train/Test) & \# Training & \# Testing & \# Classes \\
\midrule
STL-10        & Train/Test & 5{,}000  & 8{,}000  & 10  \\
CIFAR-10      & Train/Test & 50{,}000 & 10{,}000 & 10  \\
CIFAR-20      & Train/Test & 50{,}000 & 10{,}000 & 20  \\
CIFAR-100     & Train/Test & 50{,}000 & 10{,}000 & 100 \\
ImageNet-10   & Train/Val  & 13{,}000 & 500      & 10  \\
ImageNet-Dogs & Train/Val  & 19{,}500 & 750      & 15  \\
\midrule
DTD           & Train+Val/Test & 3{,}760 & 1{,}880 & 47 \\
UCF-101       & Train/Val      & 9{,}537 & 3{,}783 & 101 \\
\bottomrule
\end{tabular}}
\end{table}

% \begin{table}[ht]
% \centering
% \footnotesize
% \setlength{\tabcolsep}{4pt}
% \begin{tabular}{l c c c c c}
% \toprule
% \textbf{Dataset} & \textbf{Split} & \textbf{\# Train} & \textbf{\# Test} & \textbf{\# Classes} \\
% \midrule
% STL-10         & Train/Test & 5,000  & 8,000  & 10 \\
% CIFAR-10       & Train/Test & 50,000 & 10,000 & 10 \\
% CIFAR-20       & Train/Test & 50,000 & 10,000 & 20 \\
% CIFAR-100      & Train/Test & 50,000 & 10,000 & 100 \\
% ImageNet-10    & Train/Val  & 13,000 & 500    & 10 \\
% ImageNet-Dogs  & Train/Val  & 19,500 & 750    & 15 \\
% \bottomrule
% \end{tabular}
% \vspace{1mm}
% \caption{Characteristics of the benchmark datasets used in our evaluation.}
% \label{tab:datasets}
% \end{table}

\subsubsection{Evaluation Metrics}
We use three widely adopted metrics Accuracy (ACC), Normalized Mutual Information (NMI), and Adjusted Rand Index (ARI) to evaluate clustering performance. Higher values of these metrics indicate better results.

\subsubsection{Implementation Details}
\label{s:Implementation Details}
In most experiments, we used the original OpenAI CLIP \citep{radford2021learning} with the ViT-B/32 backbone \citep{dosovitskiy2020image}. For a fair comparison with TEMI \citep{adaloglou2023exploring} and CPP \citep{chu2024image}, we replaced the CLIP backbone with ViT-L/14 to match the architectures used in those methods. SCP-DINO employs the ViT-B/8 architecture from original DINO \citep{caron2021emerging}, while SCP-MIX adopts concatenated features from both CLIP and DINO. The clustering head $g$ is a five-layer MLP with the following architecture: $D \rightarrow 1024 \rightarrow 786 \rightarrow 512 \rightarrow 1024 \rightarrow K$, where $D$ is the output dimension of the pre-trained model and $K$ is the number of clusters.

We train the model using the Adam optimizer with a cosine annealing learning rate schedule, starting from an initial rate of \(1 \times 10^{-3}\) for 30 epochs. The batch size is set to 512 for all datasets. For regularization, we set $\alpha = 2$ for CIFAR-20 and ImageNet-Dogs, $\alpha = 3$ for CIFAR-100 to account for its larger number of clusters, and $\alpha = 1$ for the remaining datasets.
 All experiments are conducted on a single NVIDIA L4 GPU. Under this setup, training SCP on CIFAR-10 takes approximately one minute, excluding data augmentation.

\subsection{Main Results}
We compare our method with state-of-the-art baselines on six widely used image clustering datasets and provide feature visualizations to demonstrate that competitive performance can be achieved with a simple deep clustering pipeline.

\subsubsection{Comparison against classical and text-based method}
We first evaluate our pipeline on five widely used datasets and compare it against 18 deep clustering baselines. Since most earlier baselines adopt ResNet-34 or ResNet-18 as their backbone, we primarily focus on comparisons with CLIP and CLIP-based methods. Specifically, CLIP (zero-shot) uses CLIP’s pre-trained image and text encoders to classify images by matching them to text prompts, while CLIP (K-means) performs clustering directly on CLIP image features using K-means. Our approach includes SCP-CLIP, which employs a ViT-B/32 backbone, consistent with other CLIP-based methods. To support our claim that SOTA clustering performance can be achieved using sufficiently informative image representations alone—and to demonstrate the flexibility of SCP—we also include SCP-DINO, which adopts the ViT-B/8 backbone. In addition, SCP-MIX further combines features from both DINO and CLIP via simple concatenation. 
\begin{table*}[ht]
\caption{%
Clustering performance of various baseline methods on multiple datasets 
(all metrics are multiplied by 100). 
We highlight the best and second-best results in \textbf{bold} and \underline{underlined}, respectively. 
CLIP (K-means), CLIP (zero-shot), SCP-CLIP, SIC, and TAC \textbf{all adopt a ViT-B/32 backbone.} SCP-DINO uses \textbf{a ViT-B/8 backbone.}
SCP-MIX adopts the concatenation of features learned by CLIP \textbf{(ViT-B/32)} and DINO \textbf{(ViT-B/8)}.
}
\centering
\footnotesize
\setlength{\tabcolsep}{2pt}
\begin{tabular}{c c ccc ccc ccc ccc ccc}
\toprule
\multirow{2}{*}{Method} & \multirow{2}{*}{Text-Free} & \multicolumn{3}{c}{STL-10} & \multicolumn{3}{c}{CIFAR-10} & \multicolumn{3}{c}{CIFAR-20} & \multicolumn{3}{c}{ImageNet-10} & \multicolumn{3}{c}{ImageNet-Dogs} \\
\cmidrule(r){3-5} \cmidrule(r){6-8} \cmidrule(r){9-11} \cmidrule(r){12-14} \cmidrule(r){15-17}
 &  & NMI & ACC & ARI & NMI & ACC & ARI & NMI & ACC & ARI & NMI & ACC & ARI & NMI & ACC & ARI \\ 
\midrule
JULE \citep{yang2016jointunsupervisedlearningdeep} 
  & \ding{51} & 18.2 & 27.7 & 16.4 & 19.2 & 27.2 & 13.8 & 10.3 & 13.7 & 3.3 & 17.5 & 30.0 & 13.8 & 5.4 & 13.8 & 2.8 \\
DEC \citep{xie2016unsuperviseddeepembeddingclustering} 
  & \ding{51} & 27.6 & 35.9 & 18.6 & 25.7 & 30.1 & 16.1 & 13.6 & 18.5 & 5.0 & 28.2 & 38.1 & 20.3 & 12.2 & 19.5 & 7.9 \\
DAC \citep{chang2017deep}
  & \ding{51} & 36.6 & 47.0 & 25.7 & 39.6 & 52.2 & 30.6 & 18.5 & 23.8 & 8.8 & 39.4 & 52.7 & 30.2 & 21.9 & 27.5 & 11.1 \\
DCCM \citep{wu2019deepcomprehensivecorrelationmining} 
  & \ding{51} & 37.6 & 48.2 & 26.2 & 49.6 & 62.3 & 40.8 & 28.5 & 32.7 & 17.3 & 60.8 & 71.0 & 55.5 & 32.1 & 38.3 & 18.2 \\
IIC \citep{ji2019invariantinformationclusteringunsupervised} 
  & \ding{51} & 49.6 & 59.6 & 39.7 & 51.3 & 61.7 & 41.1 & 22.5 & 25.7 & 11.7 & - & - & - & - & - & - \\
PICA \citep{huang2020deep} 
  & \ding{51} & 61.1 & 71.3 & 53.1 & 59.1 & 69.6 & 51.2 & 31.0 & 33.7 & 17.1 & 80.2 & 87.0 & 76.1 & 35.2 & 35.3 & 20.1 \\
CC \citep{li2021contrastive} 
  & \ding{51} & 76.4 & 85.0 & 72.6 & 70.5 & 79.0 & 63.7 & 43.1 & 42.9 & 26.6 & 85.9 & 89.3 & 82.2 & 44.5 & 42.9 & 27.4 \\
IDFD \citep{tao2021clusteringfriendlyrepresentationlearninginstance} 
  & \ding{51} & 64.3 & 75.6 & 57.5 & 71.1 & 81.5 & 66.3 & 42.6 & 42.5 & 26.4 & 89.8 & 95.4 & 90.1 & 54.6 & 59.1 & 41.3 \\
SCAN \citep{van2020scan}
  & \ding{51} & 69.8 & 80.9 & 64.6 & 79.7 & 88.3 & 77.2 & 48.6 & 50.7 & 33.3 & - & - & - & 61.2 & 59.3 & 45.7 \\
MiCE \citep{tsai2020mice} 
  & \ding{51} & 63.5 & 75.2 & 57.5 & 73.7 & 83.5 & 69.8 & 43.6 & 44.0 & 28.0 & - & - & - & 42.3 & 43.9 & 28.6 \\
GCC \citep{zhong2021graph} 
  & \ding{51} & 68.4 & 78.8 & 63.1 & 76.4 & 85.6 & 72.8 & 47.2 & 47.2 & 30.5 & 84.2 & 90.1 & 82.2 & 49.0 & 52.6 & 36.2 \\
NNM \citep{dang2021nearest} 
  & \ding{51} & 66.3 & 76.8 & 59.6 & 73.7 & 83.7 & 69.4 & 48.0 & 45.9 & 30.2 & - & - & - & 60.4 & 58.6 & 44.9 \\
TCC \citep{shen2021you}
  & \ding{51} & 73.2 & 81.4 & 68.9 & 79.0 & 90.6 & 73.3 & 47.9 & 49.1 & 31.2 & 84.8 & 89.7 & 82.5 & 55.4 & 59.5 & 41.7 \\
SPICE \citep{niu2022spice}
  & \ding{51} & 81.7 & 90.8 & 81.2 & 73.4 & 83.8 & 70.5 & 44.8 & 46.8 & 29.4 & 82.8 & 92.1 & 83.6 & 57.2 & 64.6 & 47.9 \\ 
\midrule
CLIP (zero-shot) 
  & & 93.9 & 97.1 & 93.7 & 80.7 & 90.0 & 79.3 & 55.3 & 58.3 & 39.8 & 95.8 & 97.6 & 94.9 & 73.5 & 72.8 & 58.2  \\
SIC \citep{cai2023semantic}
  & & 95.3 & 98.1 & 95.9 & 84.7 & 92.6 & 84.4 & 59.3 & 58.3 & 43.9 & 97.0 & 98.2 & 96.1 & 69.0 & 69.7 & 55.8  \\
TAC \citep{li2023image}
  & & 95.5 & 98.2 & 96.1 & 83.3 & 91.9 & 83.1 & \underline{61.1} & \textbf{60.7} & \underline{44.8} & \textbf{98.5} & \underline{99.2} & \underline{98.3} & \underline{80.6} & \underline{83.0} & \underline{72.2} \\
\midrule
CLIP (K-means) 
  & \ding{51} & 92.2 & 94.5 & 89.5 & 71.3 & 75.2 & 62.6 & 50.8 & 48.1 & 30.6 & 96.9 & 98.2 & 96.1 & 39.8 & 38.1 & 20.1  \\
SCP-CLIP 
  & \ding{51} & 94.6 & 97.9 & 95.1 & 85.5 & 93.3 & 85.7 & 59.5 & 60.1 & 44.3 & 97.4 & 98.5 & 97.9 & 56.3 & 59.6 & 44.6  \\
SCP-DINO
  & \ding{51} & \underline{95.8} & \underline{98.4} & \underline{96.5} & \underline{89.2} & \underline{95.2} & \underline{89.8} & 58.9 & 59.3 & 42.8 & 95.9 & 98.8 & 96.7 & 77.1 & 80.5 & 70.5 \\
SCP-MIX 
  & \ding{51} & \textbf{97.9} & \textbf{98.9} & \textbf{97.6} & \textbf{91.7} & \textbf{96.5} & \textbf{92.5} & \textbf{62.7} & \underline{60.6} & \textbf{45.3} & \underline{98.0} & \textbf{99.3} & \textbf{98.5} & \textbf{80.6} & \textbf{84.0} & \textbf{74.5} \\
\bottomrule
\end{tabular}
\label{tab:performance_comparison_textfree}
\end{table*}

% \begin{figure*}[h]
% \centering
% \subfigure[Step 1, $ARI=4.5$]{
%     \centering
%     \includegraphics[width=0.21\textwidth]{images/step1.png}
%     \label{fig:step1}
% }
% \hfill
% \subfigure[Step 10, $ARI=24$]{
%     \centering
%     \includegraphics[width=0.21\textwidth]{images/step10.png}
%     \label{fig:step10}
% }
% \hfill
% \subfigure[Step 50, $ARI=57$]{
%     \centering
%     \includegraphics[width=0.21\textwidth]{images/step50.png}
%     \label{fig:step50}
% }
% \hfill
% \subfigure[Step 100, $ARI=59$]{
%     \centering
%     \includegraphics[width=0.21\textwidth]{images/step100.png}
%     \label{fig:step100}
% }
% \caption{%
% Visualization of representations from different training steps learned by SCP-DINO on the ImageNet-Dogs training set, 
% along with the corresponding clustering ARI (multiplied by 100). 
% (a) Image embeddings directly from the DINO image encoder, with clusters obtained using K-means. 
% (b)--(d) Image logits learned by SCP-DINO across different steps. 
% There are $77$ steps per epoch.
% }
% \label{fig:epoch_comparison_tSNE}
% \vskip -0.1in
% \end{figure*}
As shown in Table~\ref{tab:performance_comparison_textfree}, our method significantly enhances clustering performance across multiple benchmark datasets. For example, on CIFAR-10, CLIP (K-means) achieves an ACC of 75.2\%, whereas SCP-CLIP improves this by 18\%, surpassing TAC by 1.3\%. On CIFAR-20, SCP-CLIP attains an ACC of 60.1\%, outperforming SIC by 1.8\%. The relatively lower performance of SCP-CLIP on ImageNet-Dogs may stem from the difficulty of capturing discriminative features in fine-grained images, particularly in the absence of aligned textual information. However, SCP-DINO, which takes advantage of a powerful vision backbone, mitigates this limitation and achieves an ACC of 80.5\%. With a slight increase in complexity, SCP-MIX further boosts performance across all benchmarks, achieving state-of-the-art accuracy on STL-10, CIFAR-10, ImageNet-10, and ImageNet-Dogs—demonstrating the potential of SCP's structure and further supporting our claim.

Notably, SCP-CLIP relies exclusively on visual features yet still surpasses CLIP zero-shot on the most benchmarks, and achieves competitive performance even when compared to TAC on STL-10, CIFAR-10, and CIFAR-20. These results underscore its effectiveness in extracting and leveraging visual representations for clustering. Furthermore, our pipeline is entirely text-free, featuring a simpler architecture with fewer hyperparameters—making it especially advantageous for adaptation to purely visual pre-trained models. 

Overall, these results demonstrate that beyond zero-shot classification and multimodal frameworks, a simple and lightweight approach can still achieve competitive clustering performance. Moreover, SCP is flexible and applicable to a variety of pre-trained models. To explore its applicability in more diverse settings, we evaluate SCP on two additional challenging datasets and three recent pre-trained models.

\subsubsection{Comparison with recent text-free methods}
\begin{table}[t]
\caption{%
Clustering performance comparison for recent text-free methods. The best and second-best results are highlighted in \textbf{bold} and \underline{underlined}, respectively. All methods use the \textbf{ViT-L/14 backbone}, and SCP-MIX adopts the concatenation of features learned by CLIP \textbf{(ViT-L/14)} and DINO \textbf{(ViT-B/8)}.
}
\label{tab:performance_comparison2}
\centering
\footnotesize
\setlength{\tabcolsep}{3pt}
\begin{tabular}{c c cc cc cc}
\toprule
\multirow{2}{*}{Method} & \multirow{2}{*}{Text-Free} 
& \multicolumn{2}{c}{CIFAR-10} 
& \multicolumn{2}{c}{CIFAR-20} 
& \multicolumn{2}{c}{CIFAR-100} \\
\cmidrule(r){3-4} \cmidrule(r){5-6} \cmidrule(r){7-8}
 &  & NMI & ACC & NMI & ACC & NMI & ACC \\
\midrule
TEMI 
  & \ding{51}
  & 92.6 & 96.9
  & 64.5 & 61.8
  & 79.9 & 73.7 \\

CPP 
  & \ding{51}
  & 93.6 & 97.4
  & \underline{72.5} & 64.2
  & \textbf{81.8} & \underline{74.0} \\

SCP-CLIP
  & \ding{51}
  & \underline{93.8} & \underline{97.5}
  & 69.2 & \underline{65.8}
  & 81.4 & \textbf{74.1} \\
  
SCP-MIX
  & \ding{51}
  & \textbf{95.0} & \textbf{98.0}
  & \textbf{72.9} & \textbf{67.5}
  & \underline{80.3} & 73.8 \\
\bottomrule
\end{tabular}
\end{table}

% \begin{table}[ht]
% \caption{%
% Clustering performance comparison for recent text-free methods. The best and second-best results are highlighted in {\first{boldface red}} and {\second{underlined in blue}}, respectively. \textbf{All methods use the ViT-L/14 backbone, and SCP-MIX adopts the concatenation (or ensemble) of features learned by CLIP (ViT-L/14) and DINO (ViT-B/8)}.
% }
% \label{tab:performance_comparison2}
% \begin{center}
% \begin{small}
% \begin{sc}
% \resizebox{0.7\columnwidth}{!}{%
% \begin{tabular}{c c c c c c c c}
% \toprule
% \multirow{2}{*}{Method} & \multirow{2}{*}{Text-Free} 
% & \multicolumn{2}{c}{CIFAR-10} 
% & \multicolumn{2}{c}{CIFAR-20} 
% & \multicolumn{2}{c}{CIFAR-100} \\
% \cmidrule(r){3-4} \cmidrule(r){5-6} \cmidrule(r){7-8}
%  &  & NMI & ACC & NMI & ACC & NMI & ACC \\
% \midrule
% TEMI \cite{adaloglou2023exploring}
%   & \ding{51}
%   & 92.6 & 96.9
%   & 64.5 & 61.8
%   & 79.9 & 73.7 \\

% CPP \cite{chu2024image}
%   & \ding{51}
%   & \second{93.6} & {97.4}
%   & \second{72.5} & {64.2}
%   & \first{81.8} & \second{74.0} \\

% SCP-CLIP
%   & \ding{51}
%   & \second{93.6} & \second{97.5}
%   & {69.2} & \second{65.8}
%   & {80.1} & \first{74.1} \\

% SCP-MIX
%   & \ding{51}
%   & \first{95.0} & \first{98.0}
%   & \first{72.9} & \first{67.5}
%   & \second{80.3} & {73.8} \\
% \bottomrule
% \end{tabular}
% }% end of resizebox
% \end{sc}
% \end{small}
% \end{center}
% \end{table}
Compared to recent unimodal competitors, under the same settings, SCP-CLIP achieves highly competitive performance across all CIFAR benchmarks with a simpler design. With additional representations, SCP-MIX further improves the results and outperforms methods such as CPP and TEMI on both CIFAR-10 and CIFAR-20. These results underscore the effectiveness and simplicity of the SCP framework. In particular, SCP achieves these results without relying on additional feature layers, support sets, self-distillation, or exponential moving average strategies during training. This simple yet effective design makes SCP accessible to the deep clustering community—serving as both a strong text-free competitor and a practical alternative when text-guided methods are not applicable.

\subsection*{Performance on Challenging Datasets}
\label{s:Challenges}
In addition to benchmark datasets, we also tested SCP on more challenging datasets that reflect more complicated image types. 
UCF-101 \citep{soomro2012ucf101dataset101human} is a video dataset focused on human actions rather than distinct objects, reflecting more dynamic and complex visual content. 
We extract one frame from each clip and split the data into training and validation sets. 
DTD \citep{cimpoi2013describingtextureswild} contains texture-centric images without outstanding central objects, offering a different type of visual structure. 
We set $\alpha = 2$ for DTD and $\alpha = 3$ for UCF-101; other settings remain the same.

\begin{table*}[t]
\caption{%
Clustering performance on DTD and UCF-101.
Best and second-best are in \textbf{bold} and \underline{underline}, respectively.
SCP-CLIP uses ViT-B/32; SCP-DINO uses ViT-B/8.
SCP-MIX concatenates CLIP (ViT-B/32) and DINO (ViT-B/8) features.}
\label{tab:dtd_ucf101_comparison}
\centering
\footnotesize
\begin{tabular}{lcc}
\toprule
\multirow{2}{*}{Method} &
DTD (NMI / ARI / ACC) &
UCF-101 (NMI / ARI / ACC) \\
\midrule
SIC \citep{cai2023semantic} & 59.6 / 45.9 / 30.5 & 81.0 / 61.9 / 53.6 \\
TAC \citep{li2023image} &
\textbf{62.1} / \underline{50.1} / \textbf{34.4} &
\textbf{82.3} / \textbf{68.7} / \textbf{60.1} \\
SCP-CLIP & 58.4 / 46.1 / 30.3 & 77.7 / 61.2 / 51.4 \\
SCP-DINO & 57.5 / 45.3 / 29.6 & 80.9 / 64.0 / 54.4 \\
SCP-MIX &
\underline{61.3} / \textbf{50.2} / \underline{34.3} &
\underline{81.3} / \underline{66.3} / \underline{56.5} \\
\midrule
CLIP (K-means) & 57.3 / 42.6 / 27.4 & 79.5 / 58.2 / 47.6 \\
CLIP (zero-shot) & 56.5 / 43.1 / 26.9 & 79.9 / 63.4 / 50.2 \\
\bottomrule
\end{tabular}
\end{table*}

We observe that SCP still achieves competitive results on both DTD and UCF-101 without text guidance, showing its capability for broader image clustering tasks when text is unavailable.

\subsection*{Applicability of SCP}
\label{s:Applicability}

To further demonstrate the applicability of SCP, we compare its performance using three additional backbones—SigLIP v2 \citep{tschannen2025siglip2multilingualvisionlanguage}, BLIP-2 \citep{li2023blip2bootstrappinglanguageimagepretraining}, and MAE \citep{he2021maskedautoencodersscalablevision}—across all benchmark and challenging datasets in Table ~\ref{tab:backbone_comparison} . We observe that SCP consistently improves clustering performance in most cases compared to the K-means baselines.

\begin{table*}[t]
\caption{%
Clustering performance on multiple datasets compared with different backbones. We compare the K-means baseline with SCP-enhanced variant. Arrows indicate improvement (\textcolor{red}{$^{\uparrow}$}) or degradation (\textcolor{blue}{$^{\downarrow}$}) over the K-means baseline.
}
\label{tab:backbone_comparison}
\centering
\begin{small}
\begin{sc}
\renewcommand{\arraystretch}{2}
\resizebox{\textwidth}{!}{%
\begin{tabular}{c 
c c c 
c c c 
c c c 
c c c 
c c c 
c c c 
c c c}
\toprule
\textbf{Method} 
& \multicolumn{3}{c}{STL-10} 
& \multicolumn{3}{c}{CIFAR-10} 
& \multicolumn{3}{c}{CIFAR-20} 
& \multicolumn{3}{c}{ImageNet-10} 
& \multicolumn{3}{c}{ImageNet-Dogs} 
& \multicolumn{3}{c}{DTD} 
& \multicolumn{3}{c}{UCF-101} \\
\cmidrule(r){2-4} \cmidrule(r){5-7} \cmidrule(r){8-10} \cmidrule(r){11-13}
\cmidrule(r){14-16} \cmidrule(r){17-19} \cmidrule(r){20-22}
& NMI & ARI & ACC & NMI & ARI & ACC & NMI & ARI & ACC 
& NMI & ARI & ACC & NMI & ARI & ACC & NMI & ARI & ACC 
& NMI & ARI & ACC \\
\midrule

SigLIPv2 (K-means) 
& 92.0 & 84.5 & 83.2 
& 82.5 & 79.0 & 70.7 
& 60.6 & 52.3 & 37.0 
& 91.1 & 79.8 & 84.7 
& 75.6 & 73.1 & 59.4 
& 64.2 & 50.9 & 37.0 
& 76.3 & 46.4 & 34.4 \\

SCP-SigLIPv2 
& 96.2\textcolor{red}{$^{\uparrow}$} & 98.2\textcolor{red}{$^{\uparrow}$} & 96.2\textcolor{red}{$^{\uparrow}$} 
& 92.6\textcolor{red}{$^{\uparrow}$} & 96.8\textcolor{red}{$^{\uparrow}$} & 93.2\textcolor{red}{$^{\uparrow}$} 
& 64.5\textcolor{red}{$^{\uparrow}$} & 62.4\textcolor{red}{$^{\uparrow}$} & 48.4\textcolor{red}{$^{\uparrow}$} 
& 98.2\textcolor{red}{$^{\uparrow}$} & 99.3\textcolor{red}{$^{\uparrow}$} & 98.5\textcolor{red}{$^{\uparrow}$} 
& 82.4\textcolor{red}{$^{\uparrow}$} & 85.4\textcolor{red}{$^{\uparrow}$} & 76.8\textcolor{red}{$^{\uparrow}$} 
& 65.4\textcolor{red}{$^{\uparrow}$} & 53.1\textcolor{red}{$^{\uparrow}$} & 38.8\textcolor{red}{$^{\uparrow}$} 
& 80.0\textcolor{red}{$^{\uparrow}$} & 67.7\textcolor{red}{$^{\uparrow}$} & 59.0\textcolor{red}{$^{\uparrow}$} \\

\midrule
BLIP-2 (K-means) 
& 86.6 & 76.8 & 73.8 
& 93.3 & 84.7 & 85.3 
& 75.1 & 67.7 & 57.2 
& 99.1 & 99.7 & 99.4 
& 73.5 & 67.4 & 57.5 
& 65.5 & 52.3 & 37.4 
& 84.8 & 74.0 & 66.6 \\

SCP-BLIP-2 
& 97.7\textcolor{red}{$^{\uparrow}$} & 98.4\textcolor{red}{$^{\uparrow}$} & 96.6\textcolor{red}{$^{\uparrow}$} 
& 97.8\textcolor{red}{$^{\uparrow}$} & 99.2\textcolor{red}{$^{\uparrow}$} & 98.3\textcolor{red}{$^{\uparrow}$} 
& 74.5\textcolor{blue}{$^{\downarrow}$} & 72.0\textcolor{red}{$^{\uparrow}$} & 60.7\textcolor{red}{$^{\uparrow}$} 
& 99.3\textcolor{red}{$^{\uparrow}$} & 99.8\textcolor{red}{$^{\uparrow}$} & 99.5\textcolor{red}{$^{\uparrow}$} 
& 85.2\textcolor{red}{$^{\uparrow}$} & 88.7\textcolor{red}{$^{\uparrow}$} & 81.6\textcolor{red}{$^{\uparrow}$} 
& 64.8\textcolor{blue}{$^{\downarrow}$} & 52.7\textcolor{red}{$^{\uparrow}$} & 38.6\textcolor{red}{$^{\uparrow}$} 
& 88.4\textcolor{red}{$^{\uparrow}$} & 75.5\textcolor{red}{$^{\uparrow}$} & 68.9\textcolor{red}{$^{\uparrow}$} \\

\midrule
MAE (K-means) 
& 44.2 & 40.2 & 30.2 
& 40.8 & 43.6 & 27.8 
& 31.6 & 26.4 & 15.6 
& 63.5 & 69.6 & 50.5 
& 3.6 & 11.8 & 1.5 
& 27.4 & 12.7 & 6.1 
& 47.1 & 26.7 & 14.7 \\

SCP-MAE 
& 53.9\textcolor{red}{$^{\uparrow}$} & 59.7\textcolor{red}{$^{\uparrow}$} & 40.7\textcolor{red}{$^{\uparrow}$} 
& 48.6\textcolor{red}{$^{\uparrow}$} & 58.4\textcolor{red}{$^{\uparrow}$} & 39.4\textcolor{red}{$^{\uparrow}$} 
& 34.4\textcolor{red}{$^{\uparrow}$} & 35.0\textcolor{red}{$^{\uparrow}$} & 19.5\textcolor{red}{$^{\uparrow}$} 
& 84.7\textcolor{red}{$^{\uparrow}$} & 91.7\textcolor{red}{$^{\uparrow}$} & 83.4\textcolor{red}{$^{\uparrow}$} 
& 11.0\textcolor{red}{$^{\uparrow}$} & 16.2\textcolor{red}{$^{\uparrow}$} & 4.3\textcolor{red}{$^{\uparrow}$} 
& 38.9\textcolor{red}{$^{\uparrow}$} & 22.9\textcolor{red}{$^{\uparrow}$} & 10.3\textcolor{red}{$^{\uparrow}$} 
& 47.6\textcolor{red}{$^{\uparrow}$} & 26.1\textcolor{blue}{$^{\downarrow}$} & 15.1\textcolor{red}{$^{\uparrow}$} \\

\bottomrule
\end{tabular}%
} % end resizebox
\end{sc}
\end{small}
\end{table*}

Particularly, in real-world applications, SCP could easily be equipped and fit to an advanced backbone or a local representation learning approach to optimize the clustering due to the simple design.

\subsection{Visualization}
% \begin{figure}[htbp]
%     \centering
%     \includegraphics[width=0.7\columnwidth]{images/SCP-main.png}
%     \caption{The visualization of clustering performance for SCP-CLIP with \textbf{ViT-B/32 backbone}. \textit{(Left)}: An example of an image-to-image search on STL-10, showing clusters produced by CLIP (Top) and SCP (Bottom); \textit{(Right)}: Visualization of clustering performance. SCP-CLIP effectively enhances CLIP's clustering performance through a cluster head.}
%     \label{fig:visualization}
% \end{figure}
\begin{figure}[t]
    \centering
    \includegraphics[width=\columnwidth]{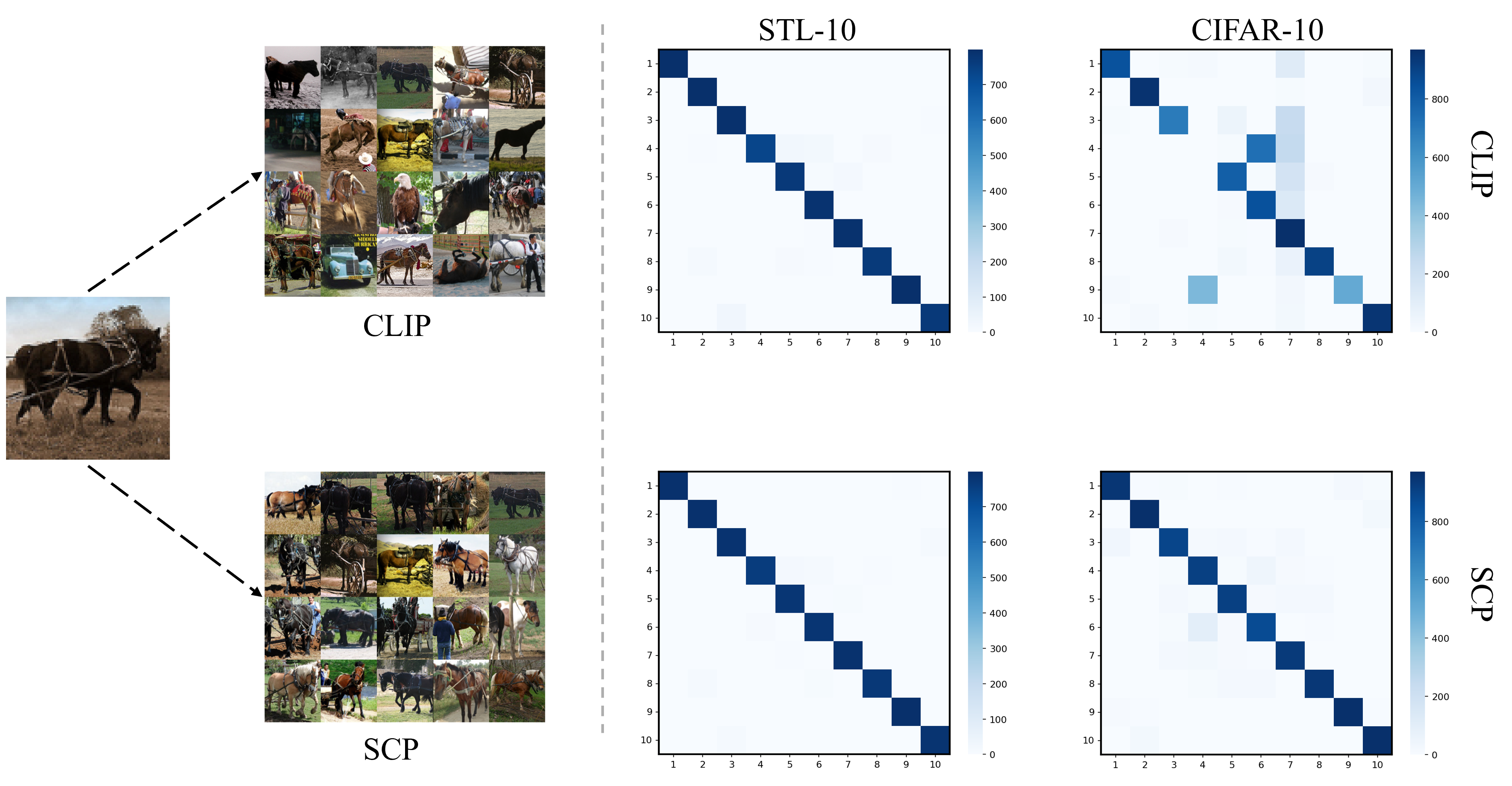}
    \caption{The visualization of clustering performance for SCP-CLIP with \textbf{ViT-B/32 backbone}. \textit{(Left)}: An example of an image-to-image search on STL-10, showing clusters produced by CLIP (Top) and SCP (Bottom); \textit{(Right)}: Visualization of clustering performance. SCP-CLIP effectively enhances original CLIP's clustering performance.}
    \label{fig:visualization}
\end{figure}
\begin{figure}[!ht]
\centering
\subfloat[DINO, $ACC=80.5\%$]{%
  \includegraphics[width=0.475\columnwidth]{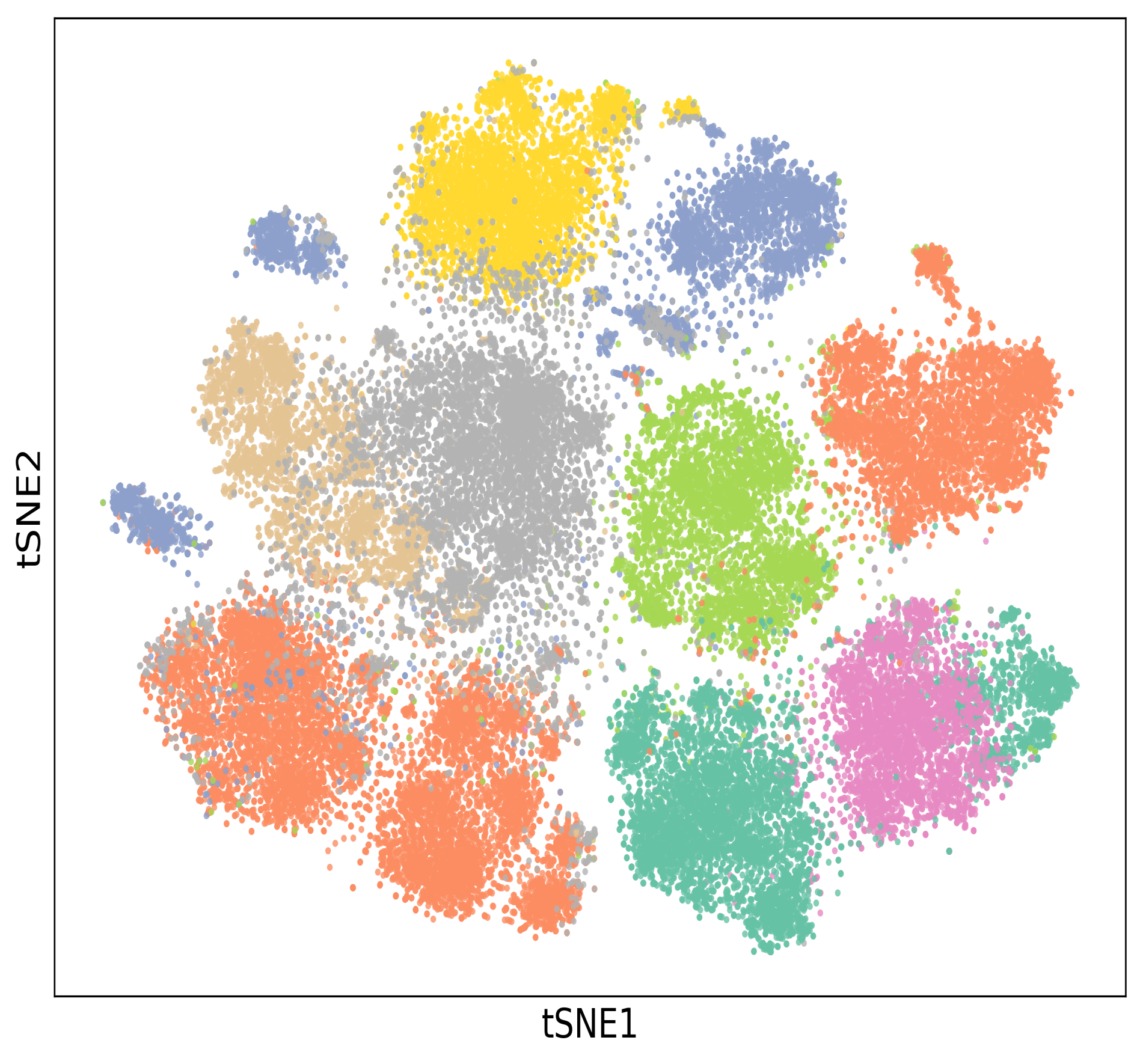}%
  \label{fig1a}}
\subfloat[CLIP, $ACC=78.5\%$]{%
  \includegraphics[width=0.475\columnwidth]{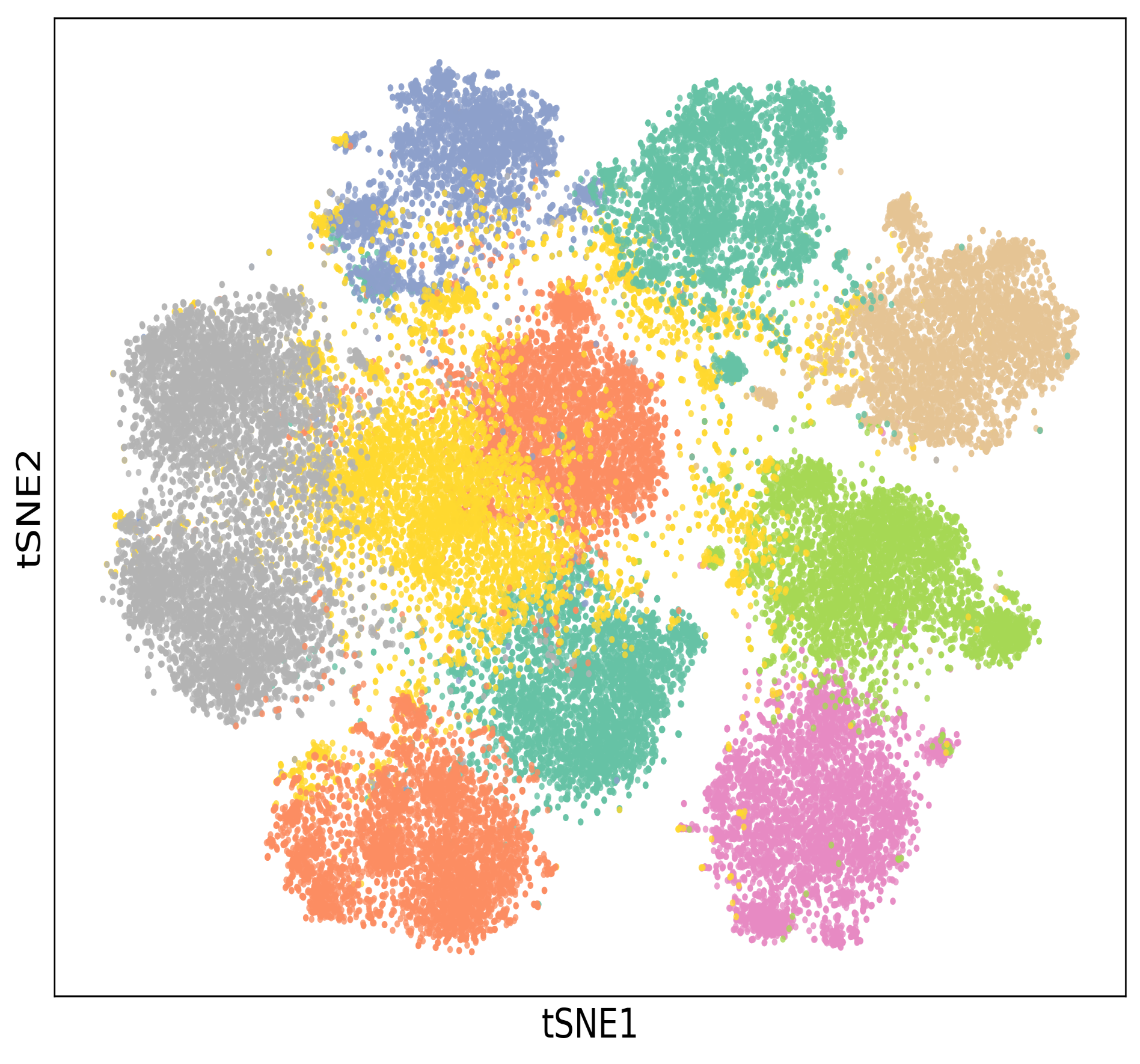}%
  \label{fig1b}}
\\[-1ex] 
\subfloat[SCP-DINO, $ACC=95.4\%$]{%
  \includegraphics[width=0.475\columnwidth]{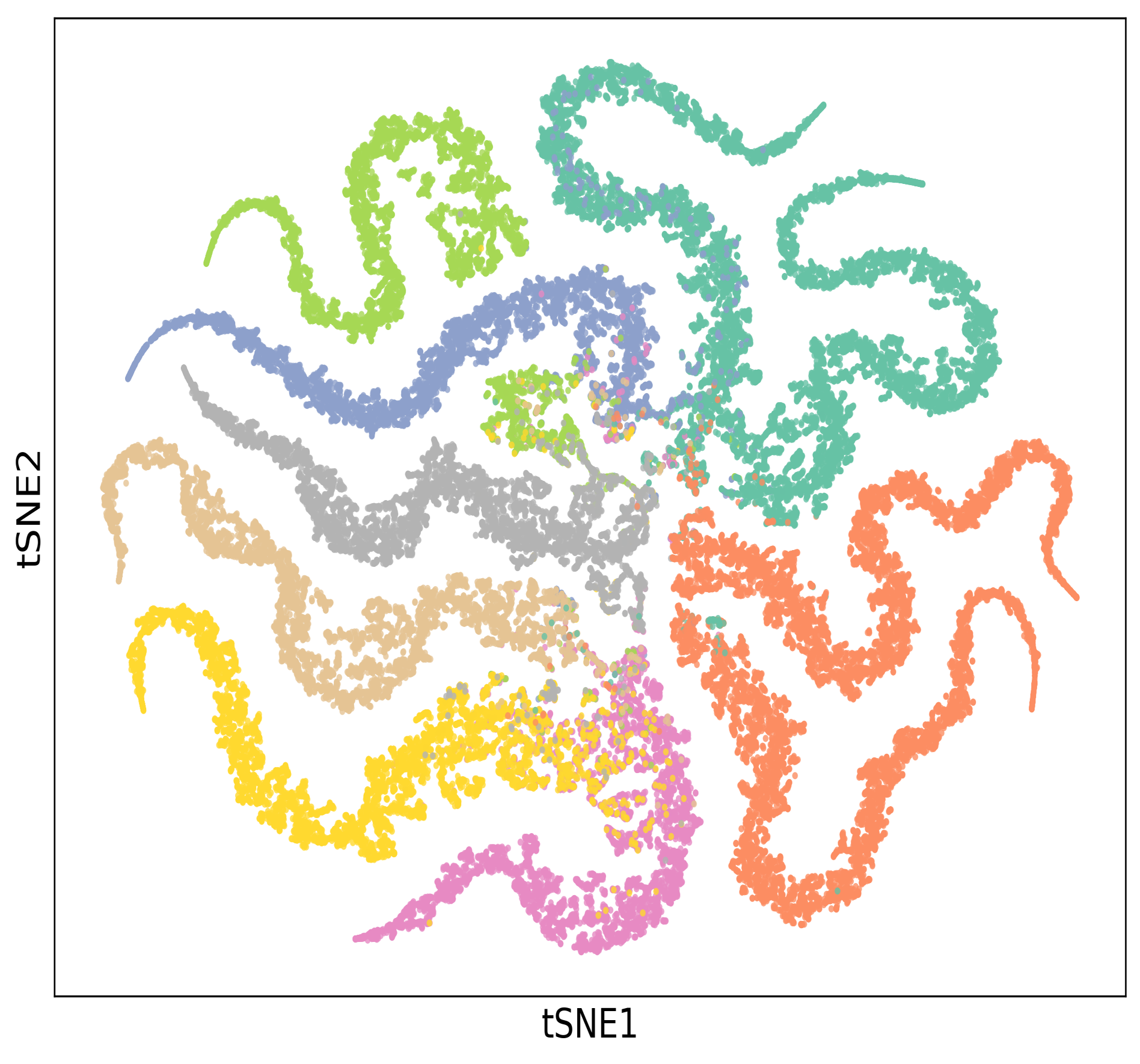}%
  \label{fig1c}}
\subfloat[SCP-CLIP, $ACC=93.5\%$]{%
  \includegraphics[width=0.475\columnwidth]{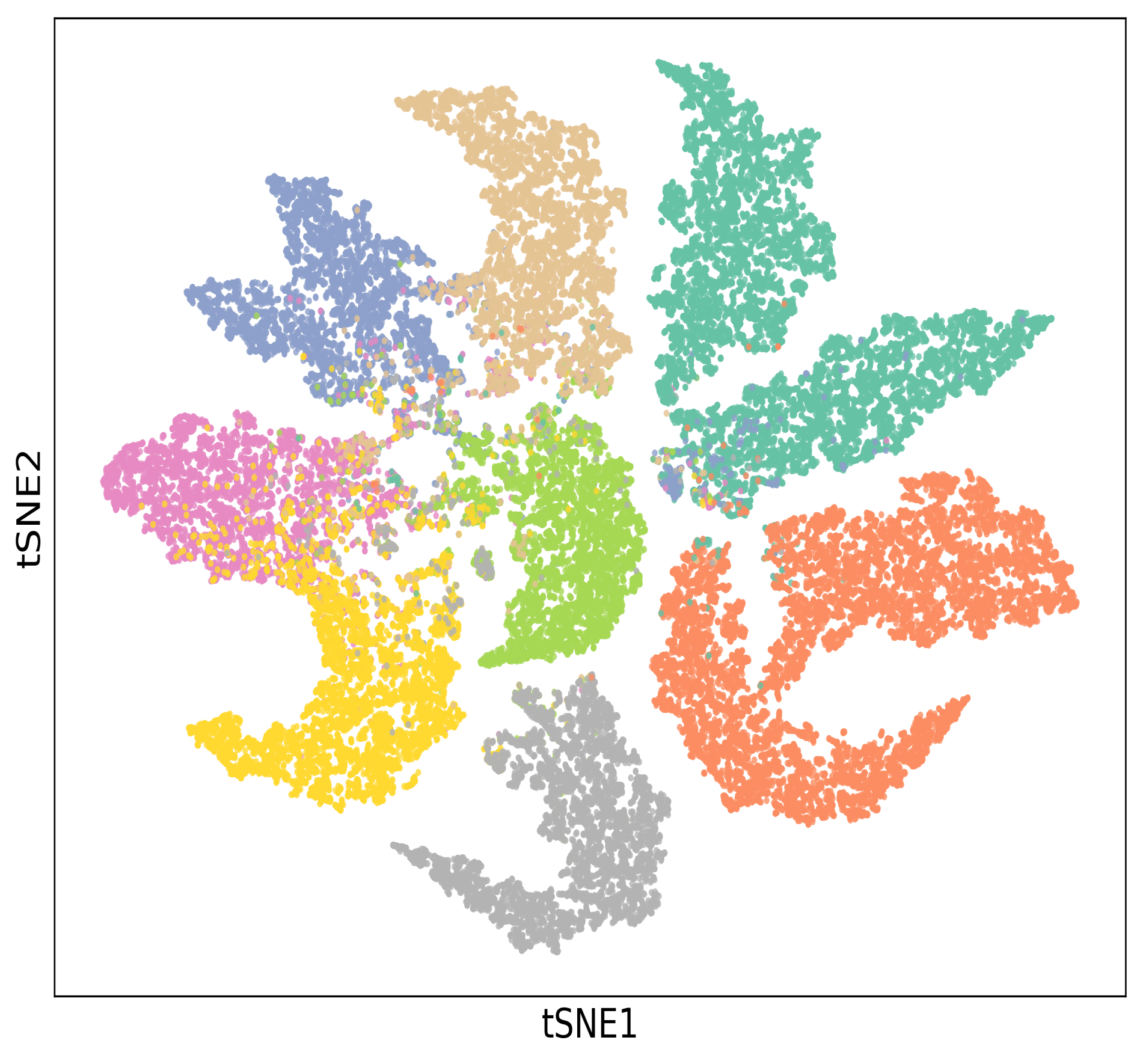}%
  \label{fig1d}}
\caption{%
Visualization of representations learned by different methods on the CIFAR-10 training set,
along with the corresponding clustering accuracy (ACC). 
(a) DINO + K-means. 
(b) CLIP + K-means. 
(c) SCP-DINO logits. 
(d) SCP-CLIP logits.}
\label{fig:clustering_comparison_tSNE}
\end{figure}
\begin{figure}[!t]
\centering
\subfloat[Step~1,~ARI~=~0.045]{%
  \includegraphics[width=0.475\columnwidth]{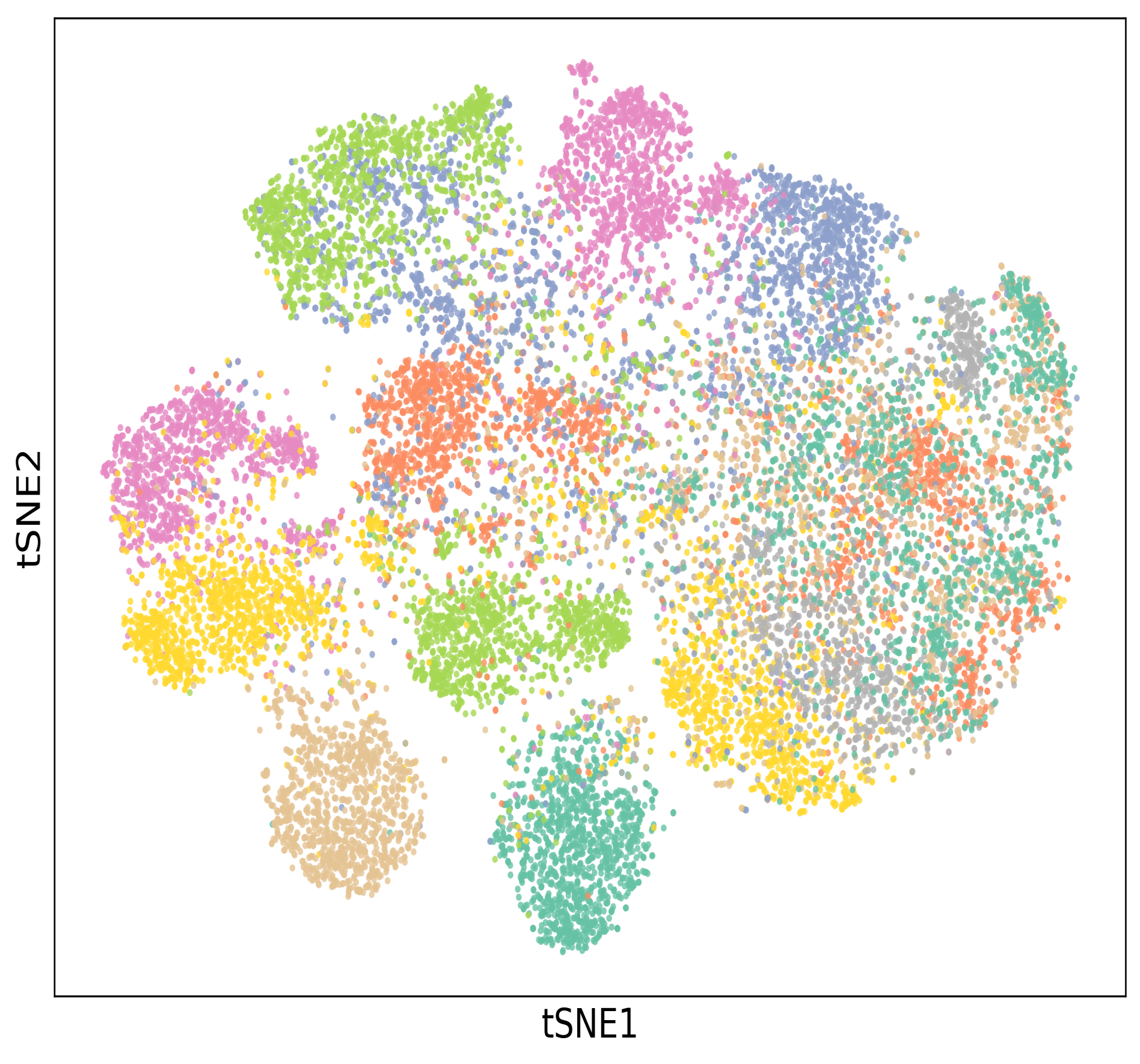}%
  \label{fig:step1}}
\subfloat[Step~10,~ARI~=~0.24]{%
  \includegraphics[width=0.475\columnwidth]{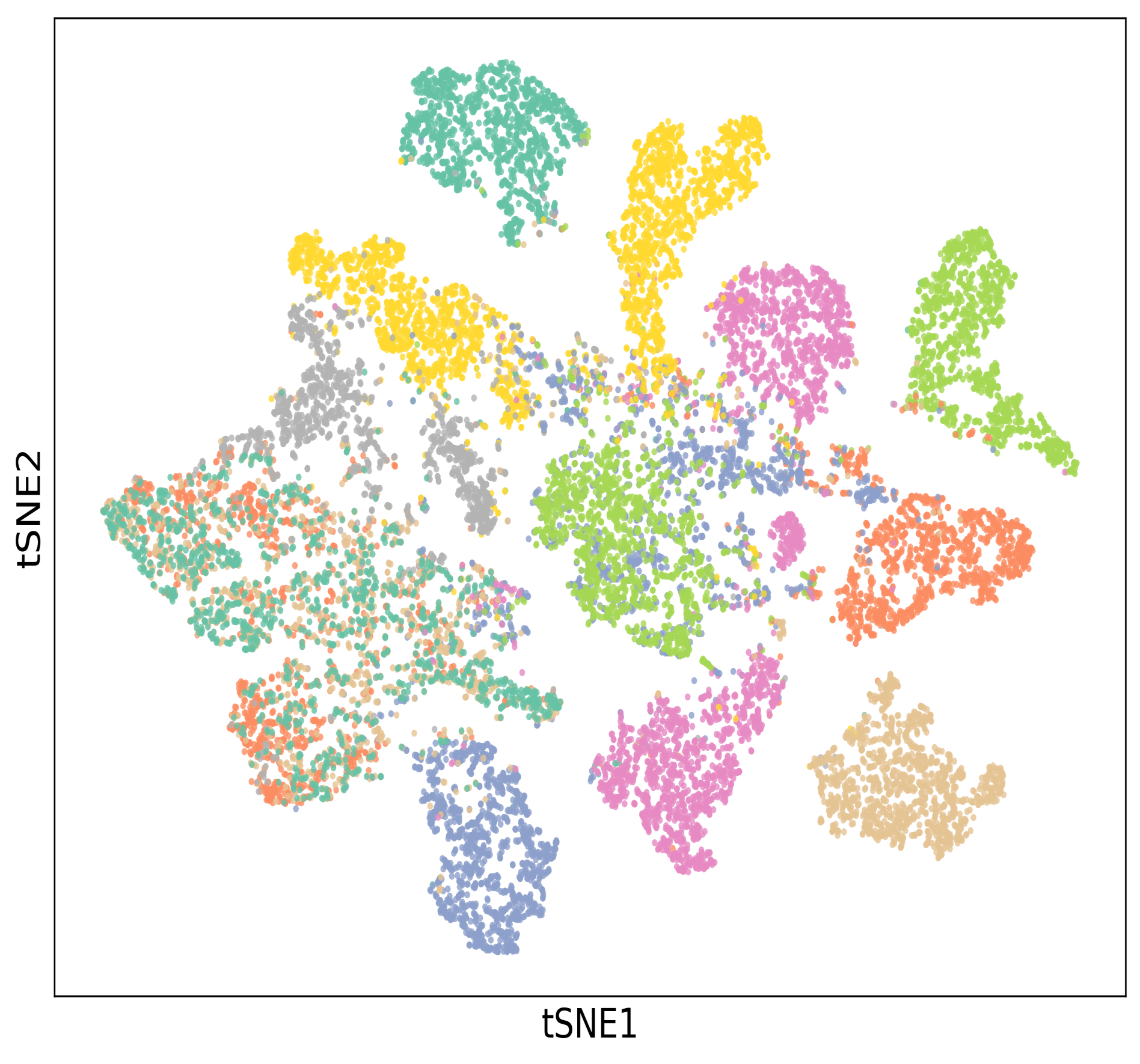}%
  \label{fig:step10}}
\\[-1ex]
\subfloat[Step~50,~ARI~=~0.57]{%
  \includegraphics[width=0.475\columnwidth]{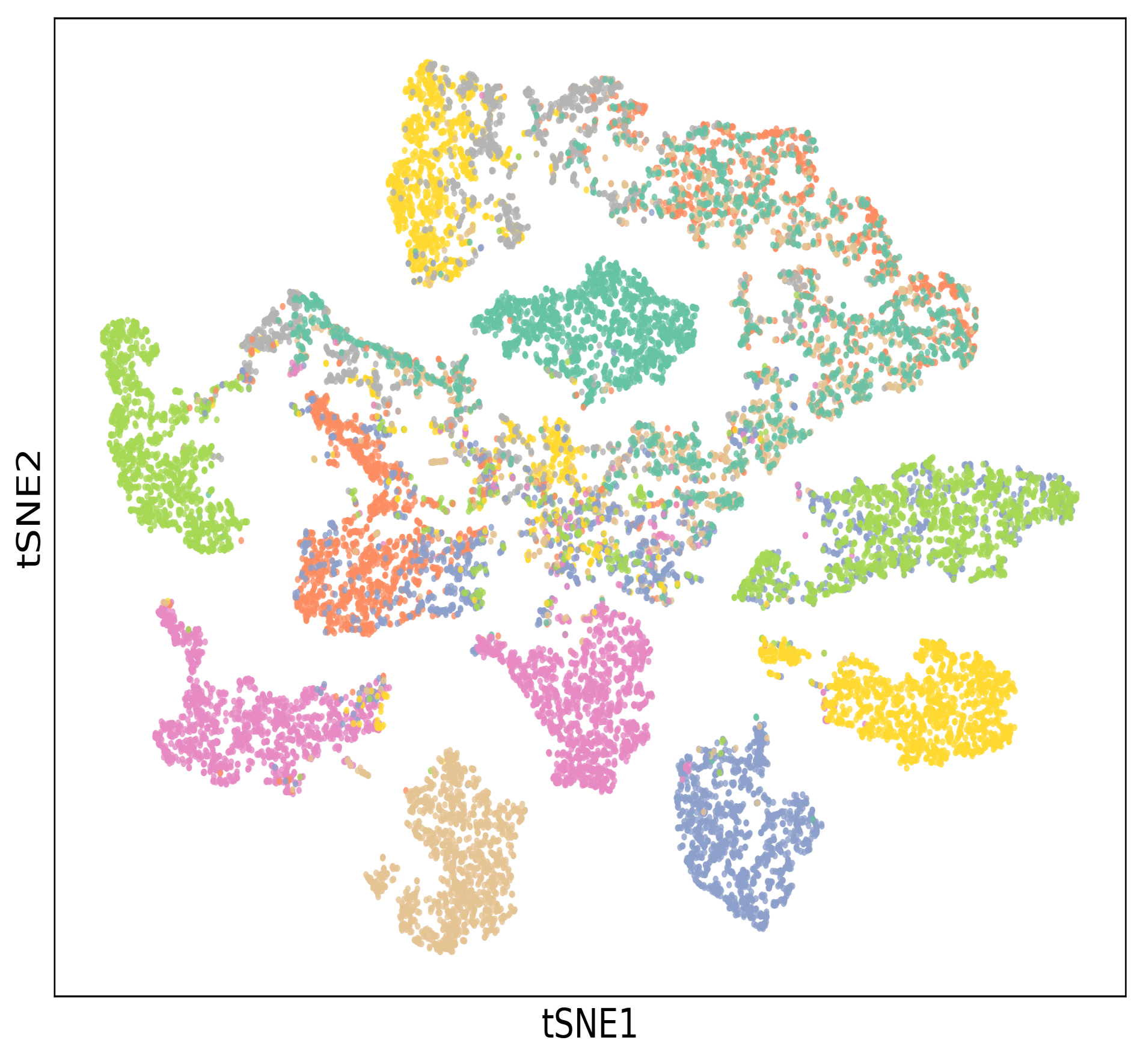}%
  \label{fig:step50}}
\subfloat[Step~100,~ARI~=~0.59]{%
  \includegraphics[width=0.475\columnwidth]{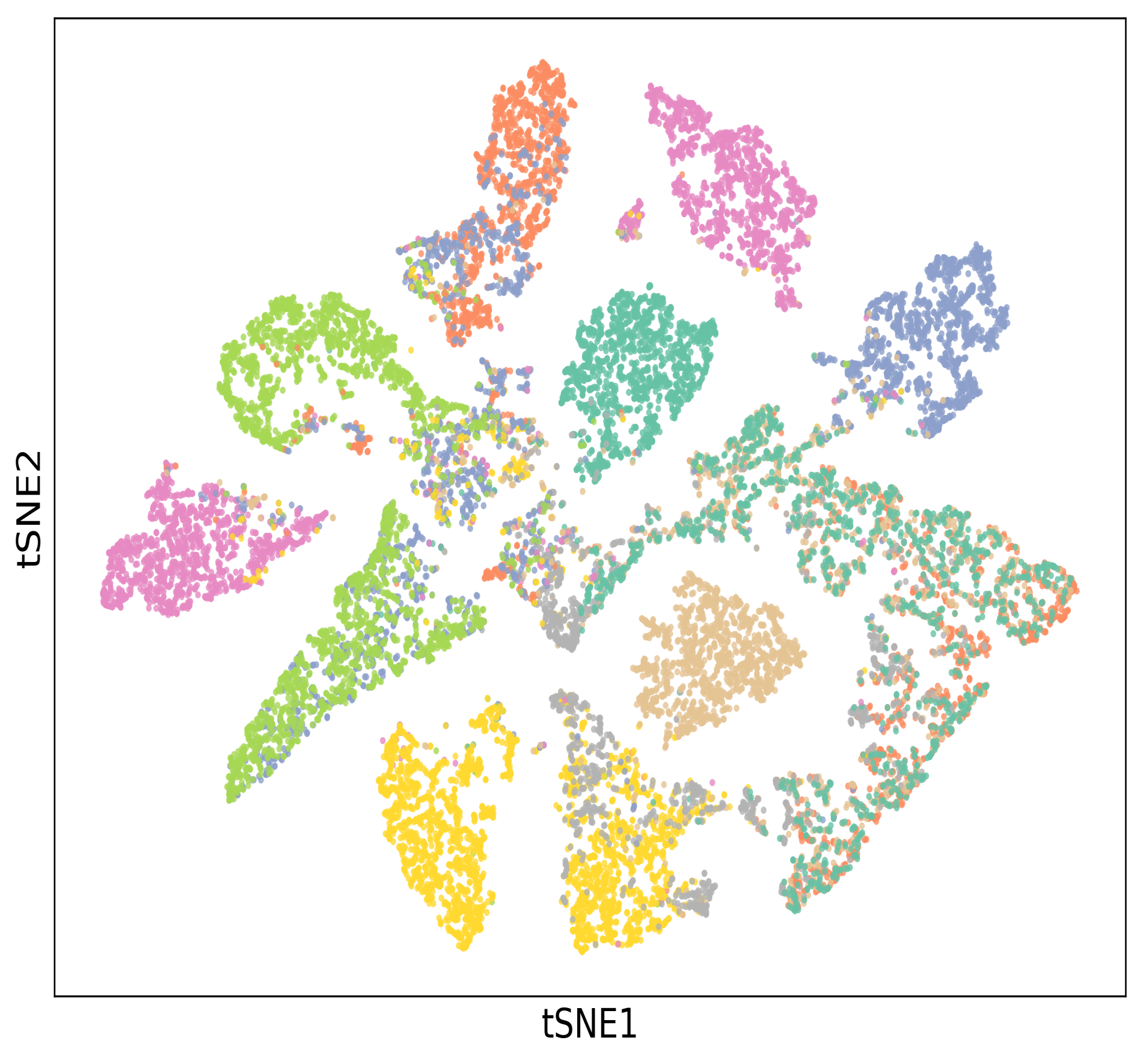}%
  \label{fig:step100}}
\caption{%
Visualization of learned representations at different training steps on ImageNet-Dogs. 
Each panel shows the t-SNE embeddings of the DINO-based features, where (a) depicts the raw encoder outputs clustered by K-means, and (b)--(d) show the logits learned by SCP-DINO across successive training stages (77 steps per epoch).}
\label{fig:epoch_comparison_tSNE}
\end{figure}

To provide an intuitive understanding of the clustering results, we visualize the clustering performance obtained from SCP in Fig. \ref{fig:visualization}, along with learned representations on the CIFAR-10 dataset in Fig. \ref{fig:clustering_comparison_tSNE}. The image logits, representing SCP outputs before the final $\operatorname{softmax}$  function, are used for visualization, with t-SNE applied to reduce the feature dimensions.  As shown, compared to embeddings directly learned by pre-trained models, SCP effectively forms well-separated clusters, leading to a higher ACC score. Without aligned text, SCP successfully extracts image features by incorporating a clustering head, resulting in superior within-cluster compactness and between-cluster separability. The visualization supports our quantitative results and highlights SCP's effectiveness in learning discriminative features suitable for clustering tasks. Furthermore, Fig. \ref{fig:epoch_comparison_tSNE} demonstrates that SCP requires only a few steps to establish a clustering structure, effectively separating certain image embeddings within 100 steps. For instance, from step 10 to step 50, SCP learns to push the light-green cluster away from the blue cluster.

\subsection{Ablation Studies}
\begin{table}[t]
\caption{%
Clustering results for different losses on CIFAR-10 and CIFAR-20. A “–” indicates model collapse.
}
\label{AS1}
\centering
\footnotesize
\setlength{\tabcolsep}{4pt}
\begin{tabular}{c c c ccc ccc}
\toprule
\multirow{2}{*}{$L_{\text{e}}$} & \multirow{2}{*}{$L_{\text{con}}$} & \multirow{2}{*}{$H(Y)$} 
& \multicolumn{3}{c}{CIFAR-10} & \multicolumn{3}{c}{CIFAR-20} \\ 
\cmidrule(r){4-6} \cmidrule(r){7-9}
 &  &  & NMI & ACC & ARI & NMI & ACC & ARI \\
\midrule
\ding{51} &      &      & –    & –    & –    & –    & –    & –    \\ 
\ding{51} & \ding{51} &      & –    & –    & –    & –    & –    & –    \\ 
\ding{51} & \ding{51} & \ding{51} & \textbf{85.5} & \textbf{93.2} & \textbf{85.7} & \textbf{59.5} & \textbf{60.1} & \textbf{44.3} \\ 
         & \ding{51} &      & –    & –    & –    & –    & –    & –    \\ 
         & \ding{51} & \ding{51} & 83.0 & 90.4 & 81.4 & \underline{55.7} & \underline{53.1} & \underline{37.6} \\ 
         &          & \ding{51} & 16.0 & 22.6 & 6.5  & 16.0 & 16.5  & 4.6  \\ 
\ding{51} &         & \ding{51} & \underline{83.9} & \underline{91.8} & \underline{82.9} & 54.7 & 52.9 & 37.0 \\ 
\bottomrule
\end{tabular}
\vspace{1mm}
\end{table}

To assess the effectiveness of the three loss terms, $L_{\text{e}}$, $L_{\text{con}}$, and $H(Y)$, we evaluate the performance of SCP using different combinations of these losses, as presented in Table \ref{AS1}. The results reveal several important insights: i) $H(Y)$ plays a crucial role in preventing cluster collapse. Without $H(Y)$, SCP tends to assign most images to only a few clusters, resulting in the model collapse on CIFAR-10 and CIFAR-20. ii) $L_{\text{con}}$ provides a slight boost in performance. The reason is that the cluster assignments would be less confident when the cluster number increases. iii) The combination of all losses effectively learns image information, leading to the best clustering performance in this table.

% \begin{table}[ht]
% \caption{%
% Clustering results for different losses on CIFAR-10 and CIFAR-20. A '-' indicates model collapse.
% }
% \label{AS1}
% \begin{center}
% \begin{small}
% \begin{sc}
% \resizebox{0.5\columnwidth}{!}{%
% \begin{tabular}{c c c c c c c c c}
% \toprule
% \multirow{2}{*}{$L_{\text{e}}$} & \multirow{2}{*}{$L_{\text{con}}$} & \multirow{2}{*}{$H(Y)$} 
% & \multicolumn{3}{c}{CIFAR-10} & \multicolumn{3}{c}{CIFAR-20} \\ 
% \cmidrule(r){4-6} \cmidrule(r){7-9}
%  &  &  & NMI & ACC & ARI & NMI & ACC & ARI \\
% \midrule
% \ding{51} &      &      & -    & -    & -    & -    & -    & -    \\ 
% \ding{51} & \ding{51} &      & -    & -    & -    & -    & -    & -    \\ 
% \ding{51} & \ding{51} & \ding{51} & \first{85.5} & \first{93.2} & \first{85.7} & \first{59.5} & \first{60.1} & \first{44.3} \\ 
%      & \ding{51} &      & -    & -    & -    & -    & -    & -    \\ 
%      & \ding{51} & \ding{51} & 83.0 & 90.4 & 81.4 & \second{55.7} & \second{53.1} & \second{37.6} \\ 
%      &      & \ding{51} & 16.0 & 22.6 & 6.5  & 16.0 & 16.5  & 4.6  \\ 
% \ding{51} &      & \ding{51} & \second{83.9} & \second{91.8} & \second{82.9} & 54.7 & 52.9 & 37.0 \\ 
% \bottomrule
% \end{tabular}
% }% end of resizebox
% \end{sc}
% \end{small}
% \end{center}
% \end{table}

\subsection{Parameter Analyses}

\begin{figure}[t]
    \centering
    \includegraphics[width=0.95\columnwidth]{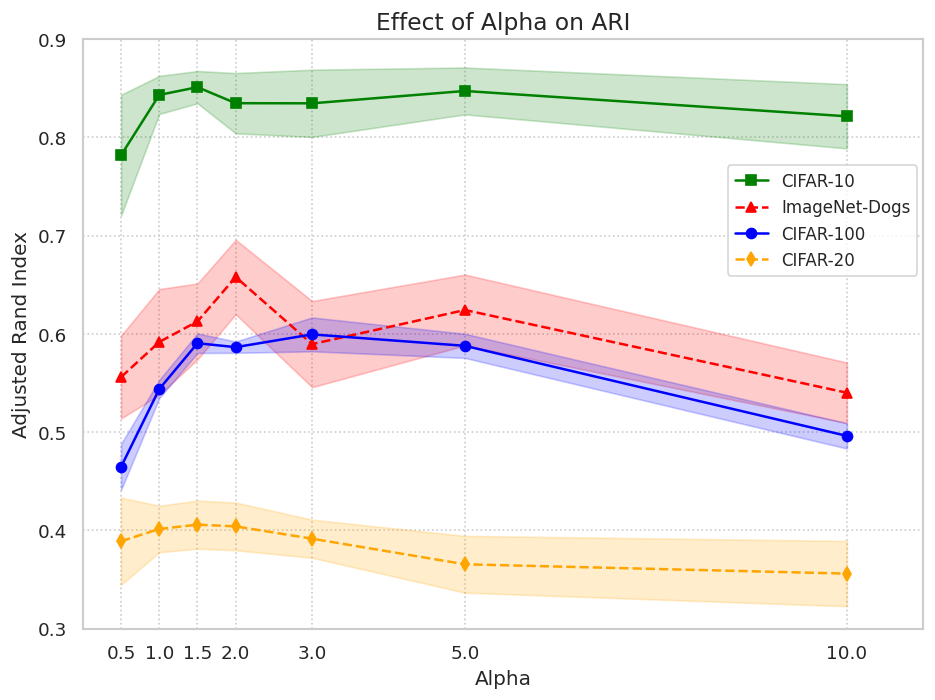}
    \caption{Comparison of different loss weights \( \alpha \). The solid line shows the mean ARI across five runs, and the shaded region indicates the standard deviation.}
    \label{fig:comparison}
\end{figure}

To show how the scale of \( H(Y) \) influences the performance of SCP, we evaluate it under various choices of \( \alpha \) on training sets of CIFAR-10, CIFAR-20, CIFAR-100 and ImageNet-Dogs under 5 random seeds. The average results and the corresponding standard deviations are presented in Fig.~\ref{fig:comparison}. We observe that setting the loss weight \(\alpha\) too low (e.g., below 0.5) makes it difficult for our method to converge. For CIFAR-10, the performance remains relatively stable and is not sensitive to large \(\alpha\) values. In contrast, for ImageNet-dogs and CIFAR-20, choosing \(\alpha = 2\) leads to better results, suggesting it provides an optimal balance between regularization and scale. For CIFAR-100, which involves a larger number of clusters, a higher \(\alpha\) value is more suitable. Based on these findings, we select \(\alpha = 3\) for CIFAR-100, \(\alpha = 2\) for CIFAR-20 and ImageNet-dogs, and \(\alpha = 1\) for all other datasets. It is also consistent with the prior knowledge of the number of clusters in each dataset.

% \begin{table}[h!]
% \centering
% \begin{tabular}{lccc}
% \toprule
% \first{Methods} & \first{CIFAR-10} & \first{CIFAR-100} & \first{STL-10} \\
% \midrule
% K-means & 83.5 & 47.3 & 94.5 \\
% NNM \cite{dang2021nearest} & 84.3 & 47.7 & 80.8 \\
% PCL \cite{li2020prototypical} & 87.4 & 52.6 & 41.0 \\
% SCAN \cite{van2020scan} & 88.3 & 50.7 & 80.9 \\
% SPICE \cite{niu2022spice} & 92.6 & 53.8 & 93.8 \\
% ProPos \cite{huang2022learning} & 94.3 & 61.4 & 86.7 \\
% \first{TEMI} \cite{adaloglou2023exploring} & 96.9 & 73.7 & 98.5 \\
% \first{CPP} \cite{chu2024image} & 97.4 & 74.0 & -- \\
% \first{CAEL} \footnotesize{(Our method)} & \first{97.2} & \first{62.4} & \first{99.6} \\
% \bottomrule
% \end{tabular}
% \vspace{0.2cm}
% \caption{\first{Clustering accuracy on CIFAR-10, CIFAR-20, and STL-10 datasets.} The K-means performance is based on features pre-processed by CLIP. TEMI (2023) \cite{adaloglou2023exploring} and CPP (2024 April) \cite{chu2024image} are the latest models that also applied CLIP, while other methods trained their own encoder. Other accuracies are collected from their original papers.}
% \label{tab:performance_comparison}
% \end{table}

\subsection*{Limitations} 
\label{s:limit}

Despite these promising results, several challenges remain. Firstly, our pipeline relies on data augmentations, which can be time-consuming when scaling to larger datasets such as ImageNet-1k. Also, our approach requires prior knowledge of the number of clusters, which can be difficult to determine in practical applications unless the practitioner possesses strong domain knowledge. Lastly, our framework's performance is bound by the quality of the pre-trained model used. However, we anticipate these to continue to improve in the future.

\section{Conclusion}
In this work, we investigate whether SOTA performance can be achieved using a simple deep clustering method. This investigation is motivated by the observation that many current methods rely on text or extra frameworks, making training complicated and widespread adoption challenging. We first provide a theoretical justification, arguing that strong performance can be achieved using informative visual representations alone. Based on this, we propose SCP, a simple and text-free adapter that requires no additional components during training, aiming for effective and accessible clustering. By leveraging pre-trained models, SCP-CLIP already demonstrates competitive performance on CIFAR-10, CIFAR-20, and STL-10. To further support our claim, we introduce SCP-DINO and SCP-MIX, which leverage stronger visual representations and achieve SOTA results across multiple benchmarks. Our findings demonstrate that SCP provides a practical and effective alternative for image clustering, featuring a simple design, easy deployment, broad accessibility, and competitive performance.

\ifCLASSOPTIONcompsoc
  \section*{Acknowledgments}
\else
  \section*{Acknowledgment}
\fi

The authors gratefully acknowledge %financial support from the Ontario Graduate Scholarship, Faculty of Science Research Fellows, and Canada Research Chairs programs.
%We appreciate 
extensive computational support from the Digital Research Alliance of Canada. %We would also like to thank Profs. Paul Ayers and Manaf Zargoush for the fruitful discussions during the preparation of this paper.
This work was supported by a Dorothy Killam Fellowship, the Canada Research Chairs program, and an NSERC Discovery Grant.

% References
%\printbibliography
\bibliographystyle{IEEEtran}
\bibliography{aaai2026}

\newpage

\end{document}